\documentclass{hpp_preprint}
\wordmarklogo{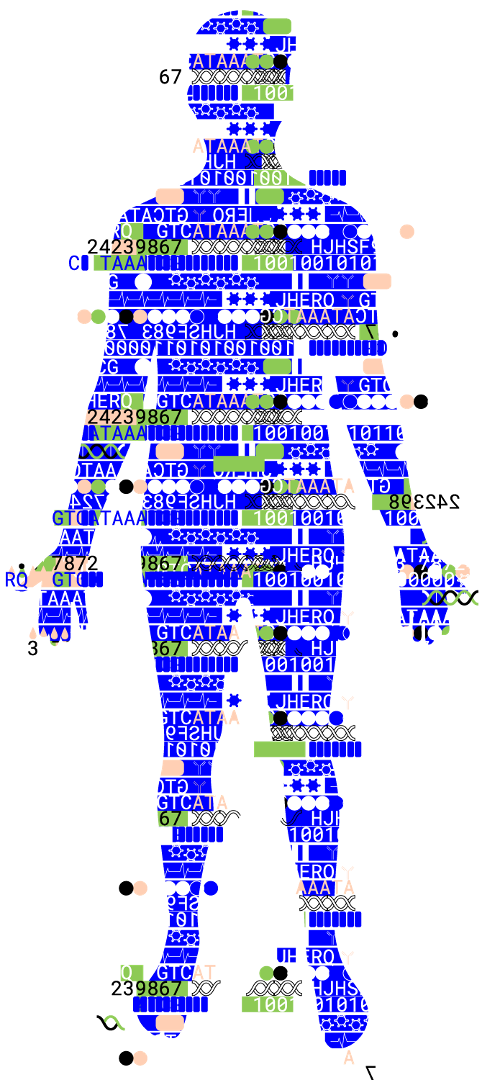}
\title{PhenoBench: Mapping What a Deeply Phenotyped Human Cohort Can Tell Us}
\shorttitle{PhenoBench}
\manuscriptauthors{Gal Sapir\textsuperscript{1,2}, Alon Diament\textsuperscript{1}, Adva Wolf\textsuperscript{1}, Doron Yaya-Stupp\textsuperscript{2}, Dikla Gelbard Solodkin\textsuperscript{1}, Dana Azouri\textsuperscript{1}, Anat Etzion-Fuchs\textsuperscript{1}, Guy Lutsker\textsuperscript{2,3}, Eran Segal\textsuperscript{2,4}, Hagai Rossman\textsuperscript{1,4,*}}
\manuscriptaffiliations{\textsuperscript{1}Pheno.AI, Tel Aviv, Israel; \textsuperscript{2}Department of Computer Science and Applied Mathematics, Weizmann Institute of Science, Rehovot, Israel; \textsuperscript{3}Department of Molecular Cell Biology, Weizmann Institute of Science, Rehovot, Israel; \textsuperscript{4}Mohamed bin Zayed University of Artificial Intelligence, Abu Dhabi, UAE}
\correspondingauthor{Hagai Rossman (hagai.rossman@mbzuai.ac.ae)}
\manuscriptkeywords{human phenotyping; multimodal evaluation; clinical prediction}
\manuscriptdate{Preprint}
\hypersetup{
  pdftitle={PhenoBench: Mapping What a Deeply Phenotyped Human Cohort Can Tell Us},
  pdfauthor={Gal Sapir et al.},
  pdfsubject={A versioned benchmark for multimodal longitudinal health data},
  pdfkeywords={human phenotyping, multimodal evaluation, clinical prediction}
}
\begin{document}
\maketitle
\begin{abstract}
Deeply phenotyped cohorts measure clinical, imaging, molecular, and wearable modalities in the same participants, combining dense observations across timescales from seconds to days with longitudinal follow-up over years. This breadth can reveal which measurements inform which health-related questions, but results from heterogeneous analyses are not directly comparable. We present PhenoBench, an executable benchmark that turns deep-phenotyping measurements into explicit questions and controlled comparisons of information sources and predictive models. It is built around the Human Phenotype Project, in which more than 13,000 participants have completed the initial visit. Each registered task specifies the target, eligibility, and timing. Each Benchmark Track defines the allowed inputs, scored split, and primary metric for an evaluation of that task. The current benchmark defines 90 clinically grounded tasks across 15 domains and 26 input modalities. As one controlled demonstration, we used PhenoBench to evaluate emerging tabular foundation models across 160 matched regression comparisons spanning 52 tasks. Under a fixed single-estimator protocol with bounded tuning, these models ranked above the evaluated task-specific baselines, including XGBoost and CatBoost, in aggregate. Giving each of the 52 tasks equal weight, the mean test-set advantage of the six pretrained models over ridge was 0.0103 \(R^2\) (95\% interval from resampling tasks with fitted results held fixed, 0.0071--0.0136). We then used the same cohort and evaluation framework to evaluate 14 language models on the validation split, collectively covering 40 task formulations spanning phenotype recovery, classification, follow-up forecasting, and participant ordering. Without cohort-specific fitting, language models made informative predictions on some tasks, but showed task-specific capability gaps, shared failures of scale, and rarely surpassed task-specific ridge or logistic regression models fitted on the same input fields. PhenoBench turns a multimodal longitudinal cohort into a versioned, auditable evaluation system where new questions, measurements, and models can be added without redefining existing comparisons.
\end{abstract}

\section{Introduction}\label{introduction}

Biomedical benchmarks typically compare models on predefined tasks within a particular data type or clinical setting. Deeply phenotyped cohorts create a different opportunity: their breadth can reveal which measurements inform which questions about human health, under what conditions, and by how much. Realizing that promise requires turning this breadth into comparable evidence across well-defined clinical and scientific questions. PhenoBench addresses this challenge using the Human Phenotype Project (HPP), translating its measurements into clinically grounded questions with explicit, comparable evaluation contracts.

The HPP is a large prospective longitudinal cohort, with more than 13,000 participants having completed the initial visit {[}19{]}. HPP combines clinical, imaging, and molecular profiling with dense physiological monitoring over days and longitudinal follow-up over years. This supports two complementary views: which measurements inform a given question, and which questions each measurement can help answer. The HPP dataset records what was measured, while individual studies estimate how informative particular measurements are for particular questions. Because those studies use different populations, inputs, splits, baselines, and methods, their findings cannot be directly compared. A common evaluation basis allows their evidence to accumulate.

PhenoBench contributes a versioned collection of 90 clinically grounded tasks across 15 domains and 26 input modalities, with human-readable Task Cards and executable evaluation contracts. These contracts support two controlled comparisons. We compare measurement routes for the same question using the same participants and fitting method, or compare methods using the same participants, inputs and split. We demonstrate these comparisons through measurement analyses, 160 matched tabular-model comparisons spanning 52 tasks, and a language-model evaluation covering 40 task formulations, including follow-up and ranking variants of registered targets. These analysis-specific sets do not cover every registered question family (Appendix~\ref{sec:task-inventory}). Together, these evaluations map where the tested measurements and methods provide predictive value and where their performance remains limited.

\subsection{Related work}\label{related-work}

Shared benchmarks offer one way to create this common evaluation basis by expressing biomedical data as fixed tasks, splits, and metrics. PhysioNet Challenges established this model around clinically motivated problems {[}22{]}, and EHRSHOT extended it to longitudinal structured EHR data by defining 15 prediction tasks for evaluating reusable representations {[}23{]}. Extending this principle to deep-phenotyping data poses a different task-definition problem. Deep-phenotyping studies collect physiological time series, imaging, clinical assays, and omics according to a research protocol, whereas EHR data record healthcare encounters, diagnoses, and interventions. These research measurements do not by themselves specify a prediction task: targets, eligible populations, temporal relations, and relevant baselines must be defined explicitly. Defining a shared set of evaluation tasks from these data is thus a distinct benchmarking problem.

Pretrained models increasingly support strong tabular and time-series prediction with little or no task-specific training {[}5--7{]}, while general-purpose language models are used for health questions at extraordinary scale: more than 300 million people ask ChatGPT health-related questions each week {[}25{]}. Evaluating these models requires specifying which question to ask, who is eligible, what information is available and when, which baseline is relevant, and how success is measured {[}3{]}. In health, these choices encode clinical judgment and determine which comparisons are meaningful and what claims the results can support.

UK Biobank prediction studies also compare methods across endpoints and predictor sets; Oexner et al.~{[}35{]}, for example, evaluated eight survival-modeling approaches across sample sizes from 5,000 to 250,000. PhenoBench complements these resources by recording the target, eligibility, timing, allowed inputs, comparator and metric as versioned contracts for research-cohort measurements. The contribution is the reusable task collection and its linked measurement and method comparisons. MedAgentBench {[}24{]} evaluates simulated EHR workflows, while MedHELM {[}31{]} evaluates a wider taxonomy of medical language-model tasks. PhenoBench instead scores cohort-grounded predictions against measured participant outcomes; these settings answer complementary questions.

\section{Results}\label{results}

\subsection{Mapping clinically grounded questions across HPP}\label{mapping-clinically-grounded-questions-across-hpp}

We registered 90 clinically grounded evaluation tasks spanning cross-modal phenotype prediction, longitudinal prediction, sequence and event prediction, retrieval and matching, and causal effect estimation. These tasks cover 15 clinical domains and 26 input modalities (Figure 1A-B), with measurement windows and task horizons spanning seconds to years (Figure 1D). Registration is distinct from evaluation coverage. Cohort sizes depend on the target and available measurement route; Appendix~\ref{sec:task-inventory} records evaluated and unevaluated entries in the released snapshots.

The initial questions were drawn from prior HPP studies and a clinician-authored inventory of clinically interpretable target measurements, then expanded as PhenoBench developed. Each question was documented in a human-readable Task Card recording the target, relevant measurements, cohort, timing, clinical context, metric, and baseline, then implemented as an executable PhenoBench task (Methods and Appendix A).

A task defines the target and who is eligible. A Benchmark Track defines an evaluation for that task by fixing the allowed inputs, scored split and primary metric. Comparisons also match the participants and baseline where required. Figure 1C shows how a question becomes a fixed comparison; Figure 2 applies these comparisons in two directions: asking which measurements inform a given question and which questions a given measurement can help answer. Within either view, holding the method fixed isolates differences among information sources, while holding the information source fixed isolates differences among methods.

\begin{figure}[!htb]
\centering
\includegraphics[width=\textwidth,height=0.70\textheight,keepaspectratio]{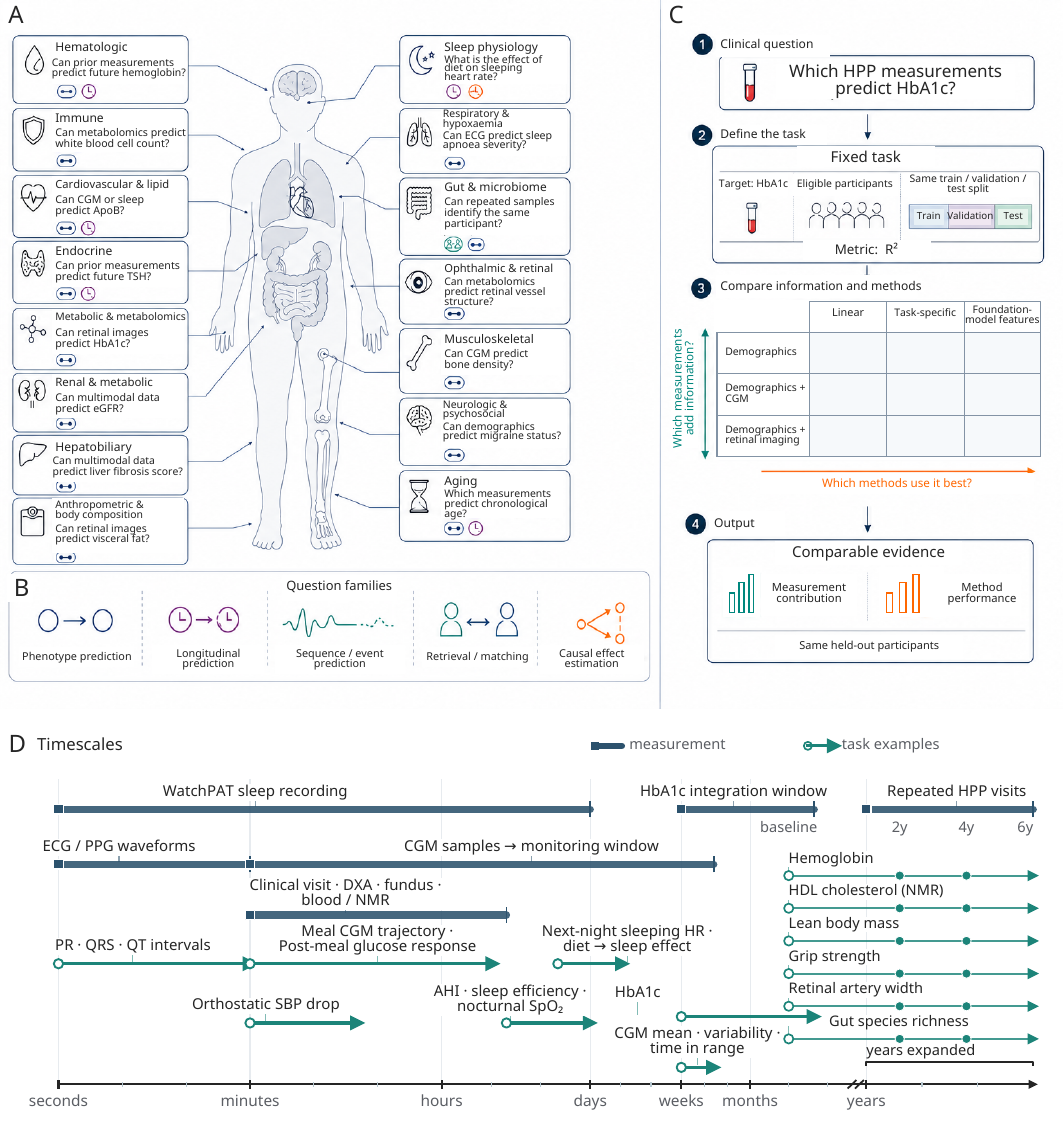}
\caption{\textbf{The HPP question space and how PhenoBench turns measurements into comparable evidence across timescales.} \textbf{A,} PhenoBench organizes 90 clinically grounded tasks across 15 clinical domains. Each domain is illustrated by a representative question; symbols indicate the task type. \textbf{B,} Questions span phenotype prediction, longitudinal prediction, sequence and event prediction, retrieval and matching, and causal effect estimation. \textbf{C,} Worked example for HbA1c prediction. The Task defines HbA1c as the target and specifies eligibility and timing. A Benchmark Track records the allowed inputs, scored split and primary metric. In this example the matched comparisons use the same participants, data split and $R^2$ metric. Input sets include demographics, continuous glucose monitoring and retinal imaging. These comparisons estimate how much predictive information is available through each prespecified measurement--representation route and how effectively different methods use the same route. \textbf{D,} Illustrative measurement windows and task horizons span seconds to years on a continuous, approximately logarithmic axis; the 1--6-year region is expanded for legibility. Task examples include baseline-to-follow-up horizons at 2, 4, and 6 years. Examples are selected rather than exhaustive, and positions communicate timescale rather than exact duration.}
\label{fig:phenobench-overview}
\end{figure}

\subsection{Mapping predictive information across tasks}\label{mapping-predictive-information-across-tasks}

Figure 2 presents PhenoBench evidence in two directions. The measurement-first view asks which questions can be predicted from a given measurement; continuous glucose monitoring (CGM) and overnight sleep monitoring (WatchPAT) are shown across several clinical domains and output types (Figure 2A). Because the examples use different cohorts and outcomes, their effect sizes should not be compared across questions. The domain distribution is shown in Supplementary Figure~\ref{fig:task-breadth-supplement}; Appendix~\ref{sec:task-inventory} indexes all 90 registered clinical tasks and their available evaluations.

In the question-first view, the question, cohort, split, metric, baseline, and ridge model are held fixed while the measurement changes (Figure 2B). Which measurement appeared most informative depended on the question: CGM had the largest validation-set \(\Delta R^2\) point estimate for fasting glucose, whereas WatchPAT had the largest for chronological age, and neither was strongest for both. Small and negative validation-set \(\Delta R^2\) values are retained; negative values indicate that adding the representation reduced performance relative to the matched baseline. The fully matched comparison across four measurement routes is limited to these two targets; the measurement-first examples do not extend its controlled coverage. Together, the two views show selective, question-dependent predictive value rather than a universal best measurement.

\begin{figure}[!htb]
\centering
\includegraphics[width=\textwidth]{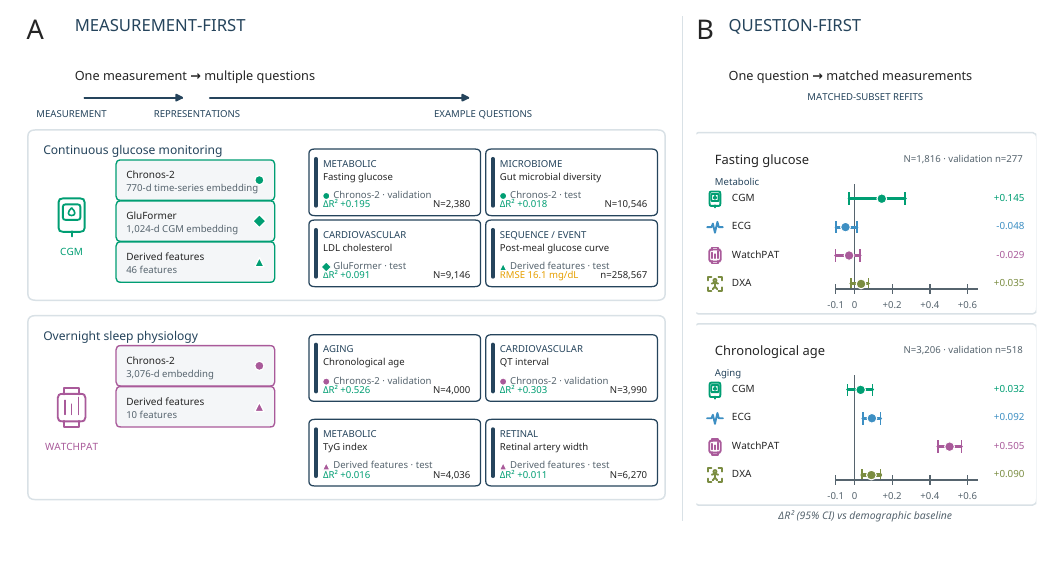}
\caption{\textbf{Two complementary readings of PhenoBench evidence.} \textbf{A,} CGM and WatchPAT are each linked through available derived or learned representations to example questions spanning metabolic, cardiovascular, microbiome, retinal, aging, and sequence outcomes. Figure 2A entries report the representation, scoring split, metric, and total cohort size. The post-meal entry reports absolute RMSE because it has no matched comparator. For that card, the displayed count is 258,567 meals across all splits. RMSE uses 39,412 test meals from that earlier evaluation. Scalar entries report $\Delta R^2$, the $R^2$ of a ridge probe with the listed representation minus that of the matched ridge probe without it, both scored on the split shown. These entries use separate cohorts, so their effect sizes should not be compared. For the GluFormer LDL cholesterol card, the test label refers to ridge scoring. The result is descriptive because its upstream CGM encoder was trained across all canonical splits. \textbf{B,} Fasting glucose and chronological age are each evaluated with CGM, electrocardiography (ECG), WatchPAT, and dual-energy X-ray absorptiometry (DXA). CGM uses frozen Chronos-2 embeddings plus glucose level and scale. ECG and WatchPAT use frozen Chronos-2 embeddings. DXA uses six curated features. Within each question, participants, split, ridge strategy, demographic comparator, and $\Delta R^2$ metric are fixed. Circles show point estimates and lines show 95\% bootstrap intervals. These matched validation refits quantify predictive association for specified measurement--representation routes, not causality or clinical utility.}
\label{fig:phenobench-two-directions}
\end{figure}

\subsection{Comparing model families on matched tasks}\label{comparing-model-families-on-matched-tasks}

\begin{figure}[!htb]
\centering
\includegraphics[width=\textwidth,height=0.68\textheight,keepaspectratio]{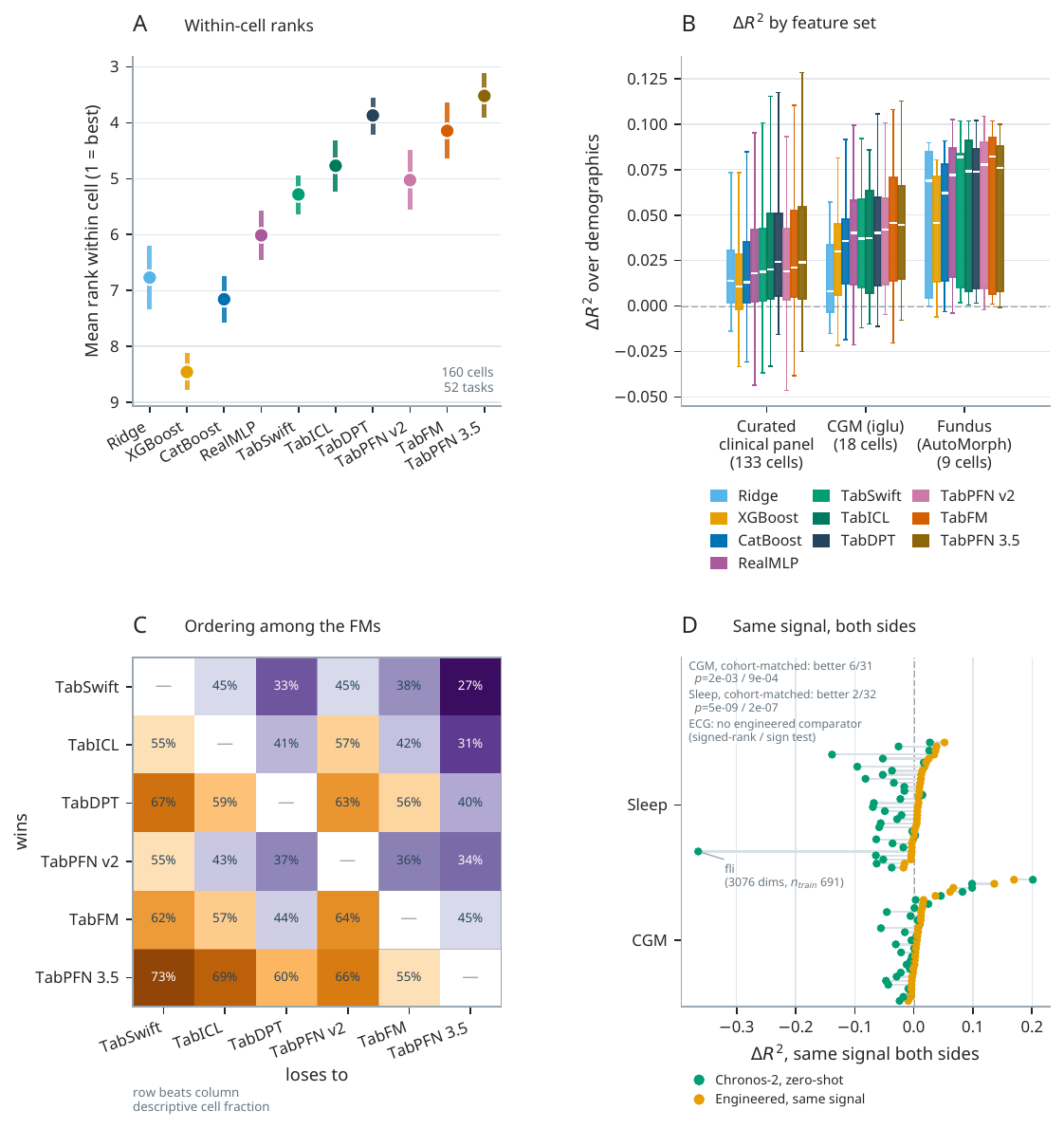}
\caption{\textbf{Model families on matched clinical comparisons.} \textbf{A,} Mean rank of each method within a comparison cell (1 = best) across 160 cells in which all ten methods ran on the same task, cohort, feature set, and split, spanning 52 tasks; bars are 95\% intervals from resampling whole tasks while retaining their cells. \textbf{B,} $\Delta R^2$ over demographics by feature set; boxes show quartiles, whiskers 1.5 IQR, outliers omitted; cell counts differ and the fundus and glucose groups are thin. \textbf{C,} Descriptive fraction of cells in which the row model exceeded the column model among the pretrained tabular models. Panels A--C summarize the finite benchmark; repeated cells within tasks are dependent. \textbf{D,} The same ridge head reads engineered features or frozen Chronos-2 representations of the \emph{same} signal. The CGM route adds level and scale to the embeddings. CGM and WatchPAT pairs match tasks and participants. WatchPAT used a post hoc intersection rerun. ECG has no engineered comparator. The labeled fatty-liver-index failure combines a 3,076-dimensional embedding with 691 training participants; the demographics-only baseline reached $R^2=0.79$. $\Delta R^2$ is measured against the same demographics-only ridge baseline within each cell.}
\label{fig:model-capacity}
\end{figure}

Across 160 matched test-set comparison cells spanning 52 tasks, the pretrained tabular models TabSwift {[}20{]}, TabICL {[}6{]}, TabDPT {[}27{]}, TabPFN v2 {[}32{]}, Google TabFM {[}33{]}, and TabPFN 3.5 {[}34{]} were compared with four task-specific baselines: ridge regression, tuned XGBoost and CatBoost, and RealMLP {[}26{]}, while holding the task, cohort, feature set, and split fixed (Figure~\ref{fig:model-capacity}). This cross-task synthesis was post hoc. Giving each of the 52 tasks equal weight, the mean advantage of the six pretrained models over ridge was 0.0103 \(R^2\) (95\% task-bootstrap interval, 0.0071--0.0136), positive in 43 tasks. On the finite benchmark of 160 task--track cells, the cell-median advantage was 0.0043 \(R^2\) (95\% task-cluster bootstrap interval, 0.0026--0.0072), and the pretrained-group mean exceeded the ridge-and-RealMLP mean in 122 of 160 cells; these cell-level summaries were descriptive because repeated tracks within a task are dependent. Task-level pairwise sensitivity retained the broad ordering. Both tuned tree libraries were below every pretrained model under the two-test correction rule, which required agreement between separately Holm-adjusted Wilcoxon signed-rank and exact sign tests, while TabPFN 3.5 was not separated from TabDPT or Google TabFM. XGBoost was below ridge in 100 of 160 cells, but the task-level contrast was unresolved; the ridge--CatBoost contrast was unresolved at both levels. These results describe the evaluated single-estimator protocol; better aggregate rank did not imply large absolute gains across all clinical comparisons.

Representations from frozen Chronos-2 {[}7{]}, a pretrained time-series model, performed worse than engineered features from the same signal in 25 of 31 CGM comparisons (\(p=0.002\)), all cohort-matched, and 30 of 32 WatchPAT comparisons rerun on exact participant intersections (\(p=4.66\times10^{-9}\)). WatchPAT pairs share exact participant cohorts. The CGM representation combines the embeddings with glucose level and scale. The intersection retained 5,574 participants globally, 97.3\% of embedding support; within each task, both arms shared the canonical split, ridge head, and penalty grid. Where Chronos-2 glucose embeddings performed better, gains were concentrated in three metabolic targets: for fasting plasma glucose, \(\Delta R^2\) was 0.201 with embeddings versus 0.169 with engineered features; for GlycA, an NMR marker of systemic inflammation, 0.099 versus 0.067; and for the triglyceride--glucose index, 0.082 versus 0.061. The gap narrowed with larger training cohorts, suggesting limited sample efficiency when fitting a ridge model to the high-dimensional embeddings.

\subsection{PhenoBench-LLM: direct prediction without cohort-specific fitting}\label{phenobench-llm-direct-prediction-without-cohort-specific-fitting}

We evaluated 14 language models using a frozen prompt and per-participant evidence packet containing age, sex, BMI, 19 CGM summary metrics, and 33 Nightingale NMR measurements. Individual models covered 22--40 of 40 task formulations spanning phenotype recovery, classification, follow-up forecasting, and participant ordering, with 150--480 validation participants per task (Figure~\ref{fig:phenobench-llm} and Methods). PhenoBench reserves this split for development. This tests direct prediction without cohort-specific fitting; the fitted comparators provide a cohort-informed reference, not a training-matched test of biomedical reasoning.

Gemini 3.7 Flash had the highest observed win rate on the eleven common tasks (common-task win rate, 82.5\%; 95\% CI, 67.1--94.4, Figure~\ref{fig:phenobench-llm}A). It led three categories, and Claude Opus 4.8 led classification. The eleven common tasks comprised six ordering, three classification, two phenotype-recovery, and no follow-up forecasting tasks. On this set, cost did not determine performance: Llama 4 Maverick, GPT-5.6, and Gemini 3.7 Flash formed the Pareto frontier (Figure~\ref{fig:phenobench-llm}B).

Task-specific fitted models generally outperformed the language models across phenotype recovery, classification, and ordering. Across phenotype-recovery tasks, language models exceeded a ridge probe fitted on the same fields in only 8 of 304 paired model--task comparisons. For follow-up tasks, the language models received each participant's baseline measurement. Relative to carrying that value forward, their gains were inconsistent, and ridge models using the same inputs performed better. The retinal artery-width task exposed a calibration mismatch hidden by aggregate rankings: the target used AutoMorph pixels, a pipeline-specific coordinate system for which the prompt supplied no mapping to the micrometre scale familiar from clinical literature, and every model answered on that micrometre-like scale. Scale-free ordering was also weak and inconsistent: full-packet Spearman \(\rho\) values were 0.109, 0.087, \(-0.040\), and \(-0.105\) across the four prespecified map models. This result therefore does not isolate biomedical reasoning from interface specification and calibration. Appendix D reports detailed estimates, probe comparisons, reliability analyses, and failure modes. These validation results are descriptive and specific to the tested tasks, packets, and prompt.

\setcounter{figure}{3}
\begin{figure}[!htb]
\centering
\includegraphics[width=\textwidth]{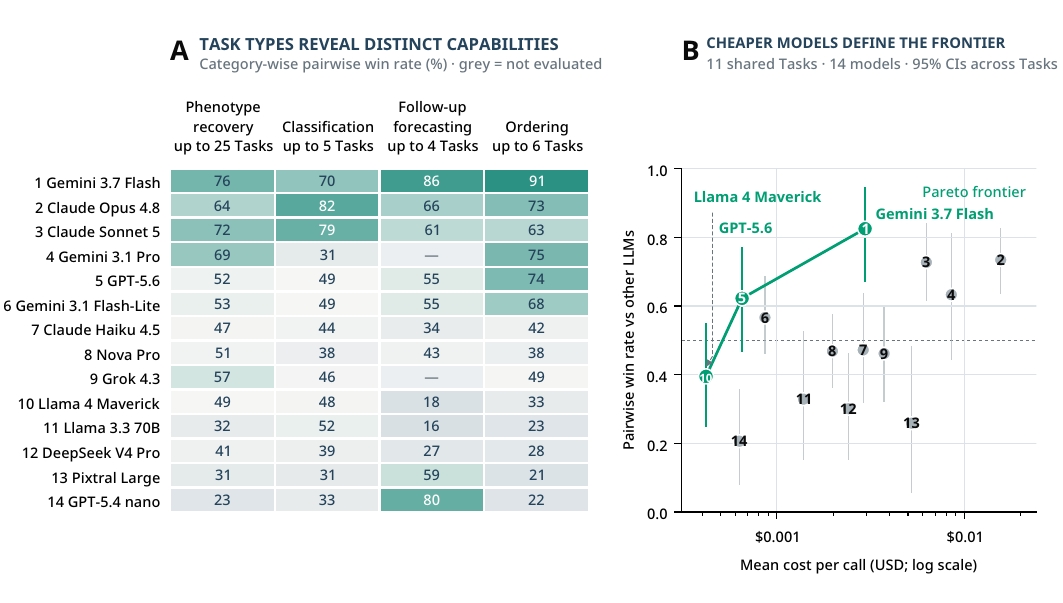}
\caption{\textbf{Cohort data maps LLM capability and efficiency.} \textbf{A,} Each cell is a model's within-category pairwise win rate against other LLMs, computed only from available head-to-heads on the same task. Each model is scored on its valid outputs, so the participants or ranking pairs can differ within a head-to-head. Values are percentages and higher is better. Longitudinal win rates compare language models with each other using raw follow-up $R^2$; they do not measure improvement over carry-forward or prediction of change. Rows are ordered by overall win rate on the 11 tasks shared by all 14 models. Grey cells indicate that a model was not evaluated in that category. Category coverage differs, and native metrics are never compared across categories. \textbf{B,} Pairwise win rate against other LLMs on the 11 shared tasks---six ordering, three classification, two phenotype-recovery, and no follow-up forecasting tasks---versus mean full-packet-arm cost per call on those same tasks. Vertical lines are 95\% bootstrap confidence intervals across tasks; numbers identify model rows in A. A model remains on the Pareto frontier when every cheaper model has a lower win rate.}
\label{fig:phenobench-llm}
\end{figure}

\section{Methods}\label{methods}

\subsection{Study design}\label{study-design}

The Human Phenotype Project (HPP) is a prospective longitudinal cohort with broad clinical, molecular, imaging, and wearable phenotyping {[}19{]}. PhenoBench analyses use task-specific eligible subsets rather than one common cohort. The default 70/15/15 train, validation, and test split keeps all observations from each participant together. We fit preprocessing on training data only. The ten-method tabular comparison uses the held-out test split. The four-route measurement comparison (Figure 2B) and PhenoBench-LLM use the validation split. Comparisons intended to isolate representation choice use the exact participant cohort shared by both representations. HPP participants provided informed consent, and identifying details were removed before computational analysis. The study was conducted according to the Declaration of Helsinki and approved by the Weizmann Institute of Science Institutional Review Board (approval 2392-4) {[}19{]}.

\subsection{Evaluations}\label{evaluations}

All Task Cards were reviewed by one physician and revised in response to their comments and corrections. The review checked each question's clinical rationale, target definition and units, timing, eligibility, comparators and leakage risks against source literature and HPP documentation (Appendix E). This is a single-reviewer check of clinical coherence, without independent specialist adjudication across all 15 domains or multi-rater validation. Cards for the earliest executable tasks were backfilled; Appendix A describes their evolution.

A model-comparison cell fixes the task, cohort, predictor, evaluation split, target, participants, and feature set while varying the fitting method. The regression capacity analysis compared ridge, tuned XGBoost and CatBoost, RealMLP, TabSwift, TabICL, TabDPT, TabPFN v2, Google TabFM, and TabPFN 3.5 in 160 complete test-set cells spanning 52 tasks. The time-series analysis compared frozen, zero-shot Chronos-2 embeddings with features engineered from the same signal under a linear ridge head. PhenoBench-LLM used fixed evidence packets and Task contracts to compare language-model predictions with demographic and fitted-probe references. Schema-faithful synthetic bundles expose the complete external-model input and output contract without releasing participant data. The tabular analysis covers the 52 tasks with at least one matched regression comparison containing all ten methods. The LLM suite covers 40 formulations of 30 targets that can be evaluated with the fixed evidence packet, including follow-up and ranking variants. Full cohort construction, model configurations, preprocessing, provenance, exclusions, and evaluation contracts are reported in the Extended Methods and Appendix~\ref{sec:task-inventory}.

\subsection{Analysis}\label{analysis}

For model-capacity comparisons, \(\Delta R^2\) was the full-model test \(R^2\) minus the common demographics-only ridge baseline from the matched ridge run. Cell-level ranks, win fractions, and the 0.0043 cell median describe the finite benchmark of 160 task--track combinations. The primary cross-task summary first averaged cells within each task, giving each of the 52 tasks equal weight, then estimated the mean advantage of the six pretrained models over ridge with a bootstrap that resampled whole tasks. Mean-rank intervals likewise resampled whole tasks while retaining all cells belonging to each sampled task; cell-level tests were descriptive. Task-level pairwise tests and equal-model-family weighting were sensitivity analyses, with tasks still sharing participants and related outcomes. Same-signal Chronos-2 comparisons were matched by task and cohort, using the exact participant cohort shared by both predictors. Detailed estimands, matching rules, resampling, permutation analyses, and statistical assumptions are reported in the Extended Methods.

\subsection{Availability and external evaluation}\label{availability-and-external-evaluation}

The public release contains the task catalog, sanitized Task Cards, aggregate results, LLM prompts and evaluation code, prediction schemas and synthetic input/output examples. To submit a predictor for evaluation, choose a Track, develop against the synthetic bundles and open a Model submission issue with a container image pinned by digest. Maintainers run accepted images offline on private inputs and validate the predictions before publishing aggregate results. The submission interface provides prediction inputs only and requires a predictor that can run without fitting on HPP data during evaluation. Evaluation is manual and capacity-limited.

Independent reproduction of the fitted-model results requires governed access to the individual-level HPP data and the benchmark execution environment. Researchers can request HPP data access through https://humanphenotypeproject.org/data-access. The data-access process and the benchmark environment are governed separately. Appendix E specifies the container interface.

\section{Discussion}\label{discussion}

``To measure is to know'' is often attributed to Lord Kelvin. What he wrote was more precise: ``when you can measure what you are speaking about, and express it in numbers, you know something about it'' {[}16{]}. He did not have deeply phenotyped cohorts in mind. Yet these cohorts become more useful when their breadth can be expressed as answers to specific questions: which measurements inform which outcomes, under what conditions, and by how much. PhenoBench provides a common basis for making those answers comparable. The same basis lets us ask how predictive information varies across measurements and how different model families use it.

HPP is particularly suited to this approach because it measures the same participants across clinical, imaging, molecular, and wearable modalities, over timescales ranging from seconds to years. By holding the clinical question and evaluation conditions fixed, PhenoBench can compare how different measurements and models use this breadth. New tabular models can therefore be evaluated across every compatible task, while language-model-based predictors can be assessed on physician-reviewed questions under the same contracts.

Across the evaluated task subsets, which measurements appeared informative depended on the question and representation. The 90-task collection allowed emerging model families to be tested across many matched clinical problems rather than on isolated benchmarks. Pretrained tabular models ranked higher overall but improved only slightly over ridge under our bounded protocol. A separate study of TabPFN on twelve binary clinical prediction tasks likewise found limited gains over optimized baselines {[}17{]}. PhenoBench extends this empirical setting through matched comparisons of models and measurement representations. Its contribution also includes an external model-submission workflow, allowing other researchers to contribute predictors and newer models to be evaluated under the same versioned task contracts. Both tuned tree libraries ranked below every pretrained tabular model under the matched protocol. Engineered glucose and sleep features often outperformed Chronos-2 embeddings of the same signals. These comparisons show where a model family helps, where gains disappear against a matched baseline, and where additional data may change the answer.

The small average tabular gains quantify predictive accuracy under the tested protocol. They do not establish a clinically meaningful improvement: we did not evaluate downstream decisions, net benefit or patient outcomes. Ridge remains a competitive reference for this task set.

Near-perfect performance on medical licensing benchmarks does not establish broad clinical capability {[}31{]}. Medical model benchmarks often begin with questions already distilled into vignettes, fixed prediction tasks, or simulated clinical workflows {[}22--24{]}. PhenoBench makes the upstream clinical and scientific work visible: defining the target, eligible population, timing, available information, comparator, and metric is central to determining what a result means. Human-readable Task Cards record the clinical rationale, and executable evaluation contracts enforce the resulting conditions. Once these are fixed, a new model can be evaluated without redefining the question, making performance differences easier to interpret.

PhenoBench is designed to grow as a versioned evidence record. New tasks, measurements, models, and results can be added while each released version remains fixed. Because every result retains its question, evaluation conditions, and provenance, later studies can determine whether a changed answer reflects stronger evidence, a different method, or a different question.

AI capability remains a ``jagged frontier'' {[}4{]}: performance on one task does not predict performance on an adjacent one. PhenoBench-LLM showed this directly across 14 models and four task categories. Models had distinct profiles across phenotype recovery, classification, follow-up forecasting, and ordering; cost did not determine performance; and task-specific fitted models generally performed better. Fixed participants, evidence, targets, and metrics made these differences interpretable and exposed shared failures of scale and target semantics. PhenoBench-LLM therefore provides a repeatable way to test how language models use cohort evidence as models, prompts, and evidence encoders change.

LLM agents can increasingly propose, execute, and revise analyses {[}18{]}. Reliable use requires a verification loop that checks each output against a defined question, held-out data, metric, comparator, and supporting evidence. The separate machine-querying experiment in Appendix C provides an initial test of whether models can navigate this evidence faithfully. With explicit rewards and protected final tasks, PhenoBench could also serve as a post-training environment for agentic research workflows {[}1,2{]}.

PhenoBench turns a deeply phenotyped cohort into a shared system for asking which measurements inform which health questions and which models can use that information. Fixing the task, cohort, available data, comparator, and metric allows new tasks and models to extend a cumulative record without redefining earlier comparisons.

\subsection{Limitations}\label{limitations}

Benchmark rankings may reify biases in HPP recruitment and measurement coverage, predictive associations may be mistaken for causal or clinically actionable evidence, and detailed health data could compromise privacy if released without controls. We mitigate these risks by retaining null results and claim boundaries, requiring matched comparisons and provenance, making cohort limitations explicit, and keeping individual-level HPP data de-identified and access controlled.

Coverage remains uneven because measurements and methods have not been evaluated on every compatible question, and cross-task summaries combine related tasks with different cohorts and outcomes. Results reflect HPP recruitment and measurement practices, so generalizability to other cohorts remains untested. Same-signal comparators are incomplete, and PhenoBench-LLM uses validation data under one prompt and evidence contract. Endpoints that did not expose a temperature parameter used provider-default decoding, an uncontrolled implementation difference across models. Reasoning-token budgets and provider configurations also differed; these are comparisons of the evaluated configurations, not equal-compute rankings. The task distribution is uneven across domains; equal task weighting does not imply equal domain weighting, and robustness to clinical-domain composition has not been assessed. Appendix F provides further details.

The tabular comparison evaluates a single-estimator CPU float32 protocol, not each model's strongest recommended ensemble. Model-specific preprocessing remains part of each implementation; Google TabFM's upstream default uses 32 estimators. External TabArena rankings provide context rather than validation of identical configurations or clinical performance. XGBoost and CatBoost used the same bounded 12-candidate tuning budget rather than exhaustive library-specific optimization. Small average gains under this protocol therefore do not settle the best attainable performance of either model family.

\section{Public benchmark leaderboards}\label{public-benchmark-leaderboards}

The \href{https://galsapir.github.io/phenobench-benchmark/}{public PhenoBench release} provides the complete task catalog, Task Cards, synthetic input/output examples, per-Track results, and a manual container-submission path; source and contribution instructions are available in the \href{https://github.com/galsapir/phenobench-benchmark}{public repository}. The compact tables below summarize the two model comparisons reported in the paper. PhenoBench does not define a universal score across heterogeneous tasks or metrics.

Appendix~\ref{sec:task-inventory} uses the \href{https://github.com/galsapir/phenobench-benchmark/blob/7b3c1af888fdb721610149835e9c6a31af150b86/data/pb/catalog.json}{v1.0.2 task catalog}. The same \href{https://github.com/galsapir/phenobench-benchmark/tree/7b3c1af888fdb721610149835e9c6a31af150b86}{benchmark snapshot} contains the public Task Cards, aggregate results and contribution instructions. It documents the \href{https://github.com/galsapir/phenobench-benchmark/blob/7b3c1af888fdb721610149835e9c6a31af150b86/data/pb/meal_cgm_earlier_evaluation.json}{earlier meal-trajectory evaluation in Figure 2A} separately from the catalog rows.

\textbf{Compact model-family leaderboard.} Mean within-cell rank across the 160 matched cells shown in Figure~\ref{fig:model-capacity}; lower is better. Rankings compare methods only inside cells with the same task, cohort, feature set, and split.

{\def\LTcaptype{none} % do not increment counter
\begin{longtable}[]{@{}lr@{}}
\toprule\noalign{}
Model & Mean rank \\
\midrule\noalign{}
\endhead
\bottomrule\noalign{}
\endlastfoot
TabPFN 3.5 & 3.52 \\
TabDPT & 3.87 \\
TabFM & 4.14 \\
TabICL & 4.77 \\
TabPFN v2 & 5.03 \\
TabSwift & 5.28 \\
RealMLP & 6.01 \\
Ridge & 6.77 \\
CatBoost & 7.16 \\
XGBoost & 8.46 \\
\end{longtable}
}

\textbf{Common-task PhenoBench-LLM leaderboard.} Pairwise win rate on the eleven Tasks evaluated by all 14 models; intervals use a Task-level bootstrap. Each comparison uses the models' valid outputs on the same task. The scored participants or ranking pairs can differ between models. This validation-set summary is descriptive.

{\def\LTcaptype{none} % do not increment counter
\begin{longtable}[]{@{}lrr@{}}
\toprule\noalign{}
Model & Win rate & 95\% CI \\
\midrule\noalign{}
\endhead
\bottomrule\noalign{}
\endlastfoot
Gemini 3.7 Flash & 82.5\% & 67.1--94.4\% \\
Claude Opus 4.8 & 73.4\% & 63.6--82.5\% \\
Claude Sonnet 5 & 72.7\% & 61.5--83.9\% \\
Gemini 3.1 Pro & 63.3\% & 44.4--81.1\% \\
GPT-5.6 & 62.2\% & 46.9--76.9\% \\
Gemini 3.1 Flash-Lite & 56.6\% & 46.2--68.5\% \\
Claude Haiku 4.5 & 47.2\% & 31.8--63.6\% \\
Nova Pro & 46.9\% & 36.4--57.3\% \\
Grok 4.3 & 46.2\% & 32.2--60.1\% \\
Llama 4 Maverick & 39.5\% & 25.2--54.9\% \\
Llama 3.3 70B & 32.9\% & 15.4--52.4\% \\
DeepSeek V4 Pro & 30.1\% & 15.4--46.2\% \\
Pixtral Large & 25.9\% & 5.6--48.3\% \\
GPT-5.4 nano & 20.6\% & 8.0--35.7\% \\
\end{longtable}
}

\section{Competing interests}\label{competing-interests}

Gal Sapir, Alon Diament, Adva Wolf, Dikla Gelbard Solodkin, Dana Azouri, Anat Etzion-Fuchs, and Hagai Rossman are employees of Pheno.AI Ltd.~Eran Segal is a paid consultant of Pheno.AI Ltd.

\section{References}\label{references}

\begin{enumerate}
\def\labelenumi{\arabic{enumi}.}
\item
  OpenAI. Learning to reason with LLMs. 2024. https://openai.com/index/learning-to-reason-with-llms/
\item
  Guo D, Yang D, Zhang H, et al.~DeepSeek-R1 incentivizes reasoning in LLMs through reinforcement learning. \emph{Nature}. 2025;645:633--638. https://doi.org/10.1038/s41586-025-09422-z
\item
  Raji ID, Bender EM, Paullada A, Denton E, Hanna A. AI and the Everything in the Whole Wide World Benchmark. 2021. https://doi.org/10.48550/arXiv.2111.15366
\item
  Dell'Acqua F, McFowland E III, Mollick E, et al.~Navigating the Jagged Technological Frontier: Field Experimental Evidence of the Effects of Artificial Intelligence on Knowledge Worker Productivity and Quality. \emph{Organization Science}. 2026;37:403--423. https://doi.org/10.1287/orsc.2025.21838
\item
  Hollmann N, Müller S, Purucker L, et al.~Accurate predictions on small data with a tabular foundation model. \emph{Nature}. 2025;637:319--326. https://doi.org/10.1038/s41586-024-08328-6
\item
  Qu J, Holzmüller D, Varoquaux G, Le Morvan M. TabICL: A Tabular Foundation Model for In-Context Learning on Large Data. 2025. https://doi.org/10.48550/arXiv.2502.05564
\item
  Ansari AF, Shchur O, Küken J, et al.~Chronos-2: From Univariate to Universal Forecasting. 2025. https://doi.org/10.48550/arXiv.2510.15821
\item
  Thompson WH, Wright J, Bissett PG, Poldrack RA. Dataset decay and the problem of sequential analyses on open datasets. \emph{eLife}. 2020;9:e53498. https://doi.org/10.7554/eLife.53498
\item
  Huffman JE. Examining the current standards for genetic discovery and replication in the era of mega-biobanks. \emph{Nature Communications}. 2018;9:5054. https://doi.org/10.1038/s41467-018-07348-x
\item
  Dwork C, Feldman V, Hardt M, et al.~The reusable holdout: Preserving validity in adaptive data analysis. \emph{Science}. 2015;349:636--638. https://doi.org/10.1126/science.aaa9375
\item
  Gibson MJ, Spiga F, Campbell A, et al.~Reporting and methodological quality of studies that use Mendelian randomisation in UK Biobank: a meta-epidemiological study. \emph{BMJ Evidence-Based Medicine}. 2023;28:103--110. https://doi.org/10.1136/bmjebm-2022-112006
\item
  Shilo S, Talmor-Barkan Y, Gorodetski M, et al.~Heterogeneity of insulin resistance surrogates in thousands of non-diabetic adults: multi-modal data reveals discordant metabolic phenotypes. \emph{medRxiv}. 2026. https://doi.org/10.64898/2026.05.02.26352290
\item
  Diament A, Gorodetski M, Jankelow A, et al.~A multimodal dataset of 21,412 recorded nights for sleep and respiratory research. \emph{arXiv}. 2023. https://doi.org/10.48550/arXiv.2311.08979
\item
  Shkolnik M, Sapir G, Shilo S, et al.~Day-to-day dietary variation shapes overnight sleep physiology: a target-trial emulation in 4.8 thousand person-nights. \emph{medRxiv}. 2026. https://doi.org/10.64898/2026.02.17.26346471
\item
  Reicher L, Bar N, Godneva A, et al.~Phenome-wide associations of human aging uncover sex-specific dynamics. \emph{Nature Aging}. 2024;4:1643--1655. https://doi.org/10.1038/s43587-024-00734-9
\item
  Thomson W. Electrical Units of Measurement. In: \emph{Popular Lectures and Addresses}. Vol. 1, Constitution of Matter. London: Macmillan and Co.; 1889:73--136. Lecture delivered May 3, 1883. https://archive.org/details/popularlecturesa01kelvuoft/page/72/mode/2up
\item
  Shaktah LA, Gustav M, Lenz T, et al.~Established machine learning matches tabular foundation models in clinical predictions. \emph{BMC Medical Informatics and Decision Making}. 2026. https://doi.org/10.1186/s12911-026-03654-3
\item
  Liu J, Pei J, Huang J, et al.~The Last Human-Written Paper: Agent-Native Research Artifacts. \emph{arXiv}. 2026. https://doi.org/10.48550/arXiv.2604.24658
\item
  Reicher L, Shilo S, Godneva A, et al.~Deep phenotyping of health--disease continuum in the Human Phenotype Project. \emph{Nature Medicine}. 2025;31:3191--3203. https://doi.org/10.1038/s41591-025-03790-9
\item
  Liu S-Y, Ye H-J. TabSwift: An Efficient Tabular Foundation Model with Row-Wise Attention. \emph{arXiv}. 2026. https://doi.org/10.48550/arXiv.2606.07345
\item
  UK AI Security Institute. Inspect AI: Framework for Large Language Model Evaluations. 2024. https://github.com/UKGovernmentBEIS/inspect\_ai
\item
  PhysioNet. George B. Moody PhysioNet Challenges. https://physionet.org/about/challenge/. Accessed September 4, 2026.
\item
  Wornow M, Thapa R, Steinberg E, Fries JA, Shah NH. EHRSHOT: An EHR Benchmark for Few-Shot Evaluation of Foundation Models. \emph{arXiv}. 2023. https://doi.org/10.48550/arXiv.2307.02028
\item
  Jiang Y, Black KC, Geng G, et al.~MedAgentBench: A Virtual EHR Environment to Benchmark Medical LLM Agents. \emph{NEJM AI}. 2025;2(9):AIdbp2500144. https://doi.org/10.1056/AIdbp2500144
\item
  OpenAI. Launching Health in ChatGPT. 2026. https://openai.com/index/health-in-chatgpt/
\item
  Holzmüller D, Grinsztajn L, Steinwart I. Better by Default: Strong Pre-Tuned MLPs and Boosted Trees on Tabular Data. \emph{arXiv}. 2024. https://doi.org/10.48550/arXiv.2407.04491
\item
  Ma J, Thomas V, Hosseinzadeh R, et al.~TabDPT: Scaling Tabular Foundation Models on Real Data. \emph{arXiv}. 2024. https://doi.org/10.48550/arXiv.2410.18164
\item
  Page EB. Ordered Hypotheses for Multiple Treatments: A Significance Test for Linear Ranks. \emph{Journal of the American Statistical Association}. 1963;58:216--230. https://doi.org/10.1080/01621459.1963.10500843
\item
  OpenRouter. OpenRouter documentation. https://openrouter.ai/docs. Accessed August 26, 2026.
\item
  Bergenstal RM, Beck RW, Close KL, et al.~Glucose Management Indicator (GMI): A New Term for Estimating A1C From Continuous Glucose Monitoring. \emph{Diabetes Care}. 2018;41:2275--2280. https://doi.org/10.2337/dc18-1581
\item
  Bedi S, Cui H, Fuentes M, et al.~MedHELM: Holistic Evaluation of Large Language Models for Medical Tasks. \emph{arXiv}. 2025. https://doi.org/10.48550/arXiv.2505.23802
\item
  Hollmann N, Müller S, Purucker L, et al.~Accurate predictions on small data with a tabular foundation model. \emph{Nature}. 2025;637:319--326. https://doi.org/10.1038/s41586-024-08328-6
\item
  Google Research. TabFM: a pretrained tabular foundation model for regression and classification. https://github.com/google-research/tabfm. Model version 1.0.0; accessed September 19, 2026.
\item
  Prior Labs. TabPFN 3.5 model card. https://huggingface.co/Prior-Labs/tabpfn\_3\_5. Accessed September 19, 2026.
\item
  Oexner RR, Schmitt R, Ahn H, et al.~Comprehensive Benchmarking of Machine Learning Methods for Risk Prediction Modelling from Large-Scale Survival Data: A UK Biobank Study. \emph{arXiv}. 2025. https://doi.org/10.48550/arXiv.2503.08870
\end{enumerate}

\appendix

\section{Task curation and ingestion audit}\label{task-curation-and-ingestion-audit}

The publication-led ingestion workflow converged on a staged, human-first review rather than a direct paper-to-code translation.

{\def\LTcaptype{none} % do not increment counter
\begin{longtable}[]{@{}
  >{\raggedright\arraybackslash}p{(\linewidth - 4\tabcolsep) * \real{0.3333}}
  >{\raggedright\arraybackslash}p{(\linewidth - 4\tabcolsep) * \real{0.3333}}
  >{\raggedright\arraybackslash}p{(\linewidth - 4\tabcolsep) * \real{0.3333}}@{}}
\toprule\noalign{}
\begin{minipage}[b]{\linewidth}\raggedright
Stage
\end{minipage} & \begin{minipage}[b]{\linewidth}\raggedright
Question answered
\end{minipage} & \begin{minipage}[b]{\linewidth}\raggedright
Review product
\end{minipage} \\
\midrule\noalign{}
\endhead
\bottomrule\noalign{}
\endlastfoot
Frame the search & Which questions can the HPP answer, and with which data? & Candidate question and relevant measurement types \\
Review HPP publications & Which clinical questions were answered or attempted in studies using the cohort? & Source-linked candidate tasks \\
Extract the task structure & What are the target, cohort, timepoint, information source, metric, comparison, and reported result? & Structured candidate record and short task name \\
Assemble clinical context & Why does the question matter, how is the target measured, and what are the caveats and meaningful comparators? & Human-readable Task Card grounded in literature and clinical review \\
Translate into PhenoBench & Does the candidate require a new Task or a new Predictor, Strategy, configuration, metric, or artifact? & Executable contract with explicit provenance \\
Review real evidence & Does the implementation run on the intended cohort, and how closely does it match the source study? & Run record, baseline comparison, comparability label, and bounded claim \\
\end{longtable}
}

A clinician-authored surrogate-target inventory also seeded clinically interpretable questions, and later benchmark work added question families not represented in the reviewed publications. These inputs broadened the collection, but the table above describes the paper-ingestion process used to turn prior HPP work into reviewable tasks.

The June 2026 corpus pass considered 61 materialized study directories. It generated 31 candidate asset families and assigned every study to one implementation queue:

{\def\LTcaptype{none} % do not increment counter
\begin{longtable}[]{@{}
  >{\raggedright\arraybackslash}p{(\linewidth - 4\tabcolsep) * \real{0.3000}}
  >{\raggedleft\arraybackslash}p{(\linewidth - 4\tabcolsep) * \real{0.4000}}
  >{\raggedright\arraybackslash}p{(\linewidth - 4\tabcolsep) * \real{0.3000}}@{}}
\toprule\noalign{}
\begin{minipage}[b]{\linewidth}\raggedright
Queue
\end{minipage} & \begin{minipage}[b]{\linewidth}\raggedleft
Studies
\end{minipage} & \begin{minipage}[b]{\linewidth}\raggedright
Interpretation
\end{minipage} \\
\midrule\noalign{}
\endhead
\bottomrule\noalign{}
\endlastfoot
Executable with existing PhenoBench components & 13 & Existing Task, Predictor, Strategy, metric, and configuration patterns were sufficient to test a paper-inspired row \\
Required a reviewed artifact & 11 & The benchmark could consume the result, but a versioned feature, embedding, or prediction artifact was missing \\
Required a reusable benchmark capability & 15 & A target shape, split policy, scorer, data route, or provenance rule was missing \\
Required manual design & 6 & The direction was plausible, but target choice or claim semantics required an author decision \\
Rejected or deferred & 16 & No selected benchmark question, duplicated coverage, protocol or reference context only, or inadequate paper evidence \\
\end{longtable}
}

The 13 executable studies were then checked against real PhenoBench runs and the original result anchors. One produced a close paper-inspired comparison, ten produced useful but non-parity analogues, and two produced weak rows. This second pass changed the ingestion criterion: a runnable configuration was no longer considered sufficient. Each applied row also required a real-data result, a relevant baseline, an explicit comparability label, and a statement of what could not be claimed.

Representative mappings illustrate why a study did not imply one new Task. The insulin-resistance study by Shilo et al.~supplied VAT questions that primarily reused an existing target while expanding information-source and configuration coverage {[}12{]}. The sleep-resource study by Diament et al.~supported a resting-heart-rate target and a sleep-derived Predictor, with its published comparison retained as an interpretation anchor {[}13{]}. The day-to-day diet and sleep study required an event-level, temporally ordered contract and artifact provenance rather than a conventional participant-level scalar row {[}14{]}. The phenome-wide aging study by Reicher et al.~reused chronological age as an evaluation target while treating system-specific age models and their outputs as versioned artifacts {[}15{]}.

\subsection{Task Card evolution and contribution path}\label{task-card-evolution-and-contribution-path}

Task Cards were introduced after the earliest executable Tasks, then backfilled as the task inventory expanded. The converged authoring process in the Methods is therefore the prospective contribution contract, not a claim that every historical Task was originally created card-first.

A Task Card provides a compact, human-readable specification before implementation details dominate the discussion:

{\def\LTcaptype{none} % do not increment counter
\begin{longtable}[]{@{}
  >{\raggedright\arraybackslash}p{(\linewidth - 2\tabcolsep) * \real{0.5000}}
  >{\raggedright\arraybackslash}p{(\linewidth - 2\tabcolsep) * \real{0.5000}}@{}}
\toprule\noalign{}
\begin{minipage}[b]{\linewidth}\raggedright
Task Card field
\end{minipage} & \begin{minipage}[b]{\linewidth}\raggedright
Content
\end{minipage} \\
\midrule\noalign{}
\endhead
\bottomrule\noalign{}
\endlastfoot
Question and target & The clinical or scientific question, endpoint definition, unit, and timing \\
HPP measurement binding & Source fields or derivation, cohort, quality control, exclusions, and evaluation grain \\
Clinical context & Why the target matters and what the literature supports \\
Evaluation plan & Metrics, demographic floor, stronger comparators, and data-split constraints \\
Risks and limits & Leakage, confounding, noise, nonclaims, and unresolved domain questions \\
Provenance & Source publications, supporting searches, review status, and linked executable task \\
\end{longtable}
}

A new contribution follows the same review path:

\begin{enumerate}
\def\labelenumi{\arabic{enumi}.}
\tightlist
\item
  State the scientific question and whether it is an exact reproduction, an adaptation, or a paper-inspired analogue.
\item
  Bind the target to a documented HPP field or derivation, including units, timing, quality control, exclusions, and evaluation grain.
\item
  Decide whether the proposal requires a new Task or only a new information source, representation, method, or configuration for an existing Task.
\item
  Create or revise the Task Card before implementation-dependent scientific choices harden.
\item
  Define the baseline and comparison set before examining held-out results.
\item
  Implement the smallest executable contract and test it on synthetic data.
\item
  Run the real-data comparison, inspect support and provenance, and record a paper-faithful, analogue, weak, or non-comparable interpretation.
\item
  Reconcile the Task Card against the implementation and original evidence, leaving unresolved questions explicit for domain review.
\end{enumerate}

Task code owns enforceable target and scoring facts. Configurations and Predictors own the available information and timepoints. Strategies and run artifacts own the fitted model and predictions. Task Cards own scientific meaning, comparator rationale, caveats, and claim boundaries. This separation allows task definitions to accumulate without embedding paper-specific extraction logic or model-specific assumptions in the target itself.

\subsection{Extended results: Mapping information value across tasks}\label{extended-results-mapping-information-value-across-tasks}

The registered collection comprises 90 clinical Tasks across 15 domains and includes continuous regression, categorical classification, sequence, causal-effect, and ranking targets (Supplementary Figure~\ref{fig:task-breadth-supplement}). Three framework-only diagnostics are excluded from these counts. The main-text examples are continuous-valued and illustrate rather than exhaustively sample this breadth.

For scalar tasks, \(\Delta R^2\) compares a ridge probe using demographics plus the listed representation with a same-run demographic-only ridge probe on the same participants and folds; demographics comprise age, sex, and BMI, except that age is omitted when chronological age is the target. In the measurement-first examples, CGM representations gave \(\Delta R^2\) estimates of 0.195 (95\% bootstrap interval, 0.090--0.285; validation) for fasting glucose, 0.091 (0.062--0.120; test) for LDL cholesterol, and 0.018 (0.001--0.037; test) for gut microbial diversity. Derived CGM features predicted two-hour post-meal trajectories with a test RMSE of 16.1 mg/dL across 39,412 meals. This is absolute forecast performance rather than an incremental information estimate because that card has no matched comparator. For sleep physiology, Chronos-2 embeddings gave estimates of 0.526 (0.472--0.575; validation) for chronological age and 0.303 (0.236--0.360; validation) for QT interval; derived features gave estimates of 0.016 (\(-0.012\) to 0.047; test) for the triglyceride--glucose (TyG) index and 0.011 (0.000--0.022; test) for retinal artery width. These examples differ in cohort, outcome, and representation and therefore describe observed routes rather than rank tasks or measurements.

Among 18 CGM-summary evaluations covering 16 distinct tasks, examples included fasting glucose (\(\Delta R^2\), 0.169; 95\% bootstrap interval, 0.084--0.247), LDL cholesterol (0.057; 0.034--0.081), total branched-chain amino acids (0.034; 0.014--0.055), and retinal artery tortuosity (\(-0.009\); \(-0.018\) to \(-0.001\)). Among 27 WatchPAT-summary evaluations covering 15 distinct tasks, examples included QT interval (0.271; 0.229--0.309), QTc interval (0.034; 0.015--0.054), retinal artery width (0.011; 0.000--0.022), and retinal artery fractal dimension at the later visit (\(-0.014\); \(-0.025\) to \(-0.004\)). These are evaluation rows rather than independent tasks because some targets appear in more than one benchmark track. The intervals quantify within-row uncertainty, are not adjusted for multiplicity, and are not used to classify rows as discoveries or null results.

Within matched question-first comparisons, fasting-glucose point estimates ranged from \(-0.048\) for ECG and \(-0.029\) for WatchPAT to 0.035 for DXA and 0.145 for CGM. For chronological age, the corresponding estimates were 0.505 for WatchPAT (95\% bootstrap interval, 0.443--0.567), 0.092 for ECG (0.044--0.139), 0.090 for DXA (0.040--0.138), and 0.032 for CGM (\(-0.037\)--0.096). These marginal intervals describe each route; they are not paired intervals for differences between routes. Comparisons are made within question only because the two questions use different cohorts.

For fasting glucose, the Chronos-2 CGM route in Figure 2A gives \(\Delta R^2=0.195\) (\(N=2{,}380\) overall, \(n=369\) scored on validation). Figure 2B gives 0.145 for that route after matching all four measurements in a cohort of \(N=1{,}816\), with \(n=277\) scored on validation. The same-signal test comparison uses a separate cohort (\(N=6{,}843\) overall, \(n=1{,}029\) scored on test). Its estimates are 0.169 for engineered summaries and 0.201 for the Chronos-2 representation.

\begingroup
\renewcommand{\figurename}{Supplementary Figure}
\setcounter{figure}{0}
\begin{figure*}[t]
\centering
\includegraphics[width=\textwidth]{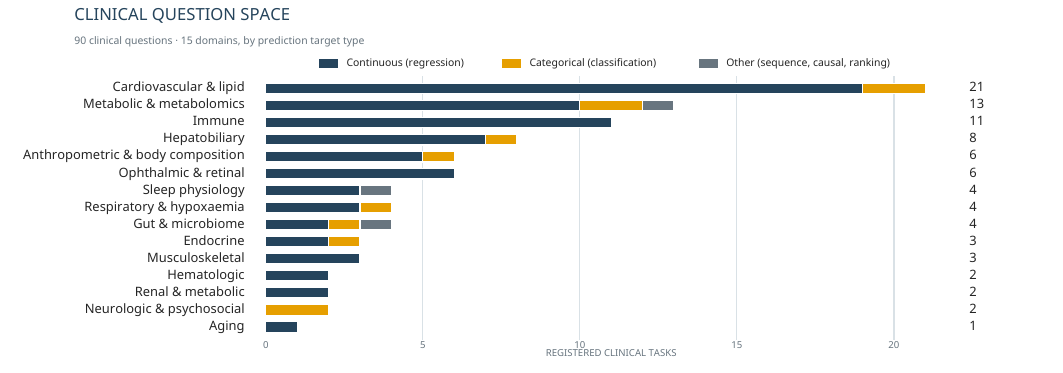}
\caption{\textbf{Registered clinical Tasks by domain and prediction target type.} PhenoBench contains 90 registered clinical Tasks across 15 clinical domains. Stacked segments denote continuous regression, categorical classification, or other sequence, causal-effect, and ranking targets. Three framework-only diagnostics are excluded. Bar length counts registered Tasks and does not represent evaluated performance.}
\label{fig:task-breadth-supplement}
\end{figure*}
\endgroup

\section{Model-capacity supporting analyses}\label{model-capacity-supporting-analyses}

\subsection{Analysis population and model roles}\label{analysis-population-and-model-roles}

Figure 3 used 160 complete test-set cells spanning 52 tasks, each with the same target, cohort, feature set, split, and all ten methods. In total, 20 cells across 10 tasks were excluded because at least one method was absent: apnea--hypopnea index, estimated glomerular filtration rate, gut Shannon diversity, gut species richness, minimum nocturnal oxygen saturation, next-night sleep, oxygen desaturation index, sleep efficiency, total sleep time, and urate. All 20 excluded cells lacked XGBoost, CatBoost, TabPFN v2, Google TabFM, and TabPFN 3.5. Eleven additionally lacked RealMLP, TabDPT, TabICL, and TabSwift; nine additionally lacked RealMLP, TabDPT, and TabICL. These are availability exclusions, not performance exclusions.

{\def\LTcaptype{none} % do not increment counter
\begin{longtable}[]{@{}
  >{\raggedright\arraybackslash}p{(\linewidth - 4\tabcolsep) * \real{0.3333}}
  >{\raggedright\arraybackslash}p{(\linewidth - 4\tabcolsep) * \real{0.3333}}
  >{\raggedright\arraybackslash}p{(\linewidth - 4\tabcolsep) * \real{0.3333}}@{}}
\toprule\noalign{}
\begin{minipage}[b]{\linewidth}\raggedright
Method
\end{minipage} & \begin{minipage}[b]{\linewidth}\raggedright
Role in Figure 3
\end{minipage} & \begin{minipage}[b]{\linewidth}\raggedright
Frozen implementation
\end{minipage} \\
\midrule\noalign{}
\endhead
\bottomrule\noalign{}
\endlastfoot
Ridge & Linear classical probe and common demographic baseline & PhenoBench persisted run configuration \\
XGBoost & Tuned tree baseline & XGBoost 3.2.0; fixed 12-candidate grid, five-fold training-set CV, CPU \\
CatBoost & Tuned tree baseline & CatBoost 1.2.10; fixed 12-candidate grid, five-fold training-set CV, CPU \\
RealMLP & Strong neural tabular baseline trained from scratch {[}26{]} & \texttt{pytabkit} 1.7.3 \\
TabSwift & Pretrained tabular model {[}20{]} & Pinned source revision and checkpoint, with full identities persisted alongside each run; 31.3 MB of checkpoint tensors, approximately 8 million parameters \\
TabICL & Pretrained tabular model {[}6{]} & \texttt{tabicl} 2.1.1; checkpoint identity persisted with each run; 108.9 MB of checkpoint tensors, approximately 29 million parameters \\
TabDPT & Pretrained tabular model {[}27{]} & \texttt{tabdpt} 1.2.0; checkpoint identity persisted with each run; 63,507,568 parameters across 647 float32 tensors (32 layers, embedding width 512, 8 heads) \\
TabPFN v2 & Pretrained regression/classification model {[}32{]} & TabPFN package 8.5.0; explicit v2 checkpoint, one estimator, CPU float32 \\
Google TabFM & Pretrained regression/classification model {[}33{]} & TabFM 1.0.0 PyTorch checkpoints; package 1.0.1; one estimator, CPU float32 \\
TabPFN 3.5 & Pretrained regression/classification model {[}34{]} & TabPFN package 9.0.0; explicit 3.5 checkpoint, one estimator, CPU float32 \\
\end{longtable}
}

The original Panel D candidate grid added 52 engineered-feature configurations: 31 for CGM and 21 for WatchPAT. The current 63 reported pairs comprise 31 CGM comparisons, selected from 21 generated and 10 pre-existing engineered runs, and 32 exact-intersection WatchPAT reruns. These configurations were retained as exploratory figure evidence rather than leaderboard rows. Panel A places the three pretrained models with directly verified parameter counts in ascending order (TabSwift, approximately 8 million; TabICL, approximately 29 million; TabDPT, 63,507,568). Comparable counts were not established for TabPFN v2, Google TabFM, and TabPFN 3.5, so their existing order was retained. The display is not interpreted as a validated total capacity ordering.

The XGBoost and CatBoost grids were fixed before their matched campaign, selected by five-fold cross-validation within the training split, and evaluated once on the held-out test split.

\subsection{Task-level and cell-level method comparisons}\label{task-level-and-cell-level-method-comparisons}

The primary cross-task summary gives each task equal weight by averaging its cells before comparing methods. In the task-level pairwise sensitivity, both tree libraries were below each pretrained model under the two-test correction rule; TabPFN 3.5, TabDPT, and Google TabFM were not separated from one another; and neither tree library was separated from ridge. The complete task-level 45-contrast family is retained in \texttt{task\_structure.json}. Tasks share participants and related outcomes, so these tests do not represent independent population replications.

As a sensitivity analysis at the descriptive cell level, Page's ordered-alternative test {[}28{]} was evaluated with ridge and RealMLP preceding all six FMs. There is no justified complete parameter-count ordering across these architectures, so the fixed display-order statistic is descriptive (Page's \(L=28{,}206\), \(p=2.50\times10^{-30}\)), not evidence of monotonic improvement with capacity. Across all 720 FM permutations, the least-favorable within-foundation-model ordering gave \(L=26{,}562\) (\(p=0.000694\)). The two tree methods were excluded from these sequences. The ordering-free within-cell mean contrast includes all six FMs. The full descriptive 45-contrast cell-level family is below; ``Robust'' records agreement between both separately Holm-adjusted tests. Exact ties are omitted from the sign-test denominator.

{\def\LTcaptype{none} % do not increment counter
\begin{longtable}[]{@{}
  >{\raggedright\arraybackslash}p{(\linewidth - 8\tabcolsep) * \real{0.1500}}
  >{\raggedleft\arraybackslash}p{(\linewidth - 8\tabcolsep) * \real{0.2000}}
  >{\raggedleft\arraybackslash}p{(\linewidth - 8\tabcolsep) * \real{0.2000}}
  >{\raggedleft\arraybackslash}p{(\linewidth - 8\tabcolsep) * \real{0.2000}}
  >{\centering\arraybackslash}p{(\linewidth - 8\tabcolsep) * \real{0.2500}}@{}}
\toprule\noalign{}
\begin{minipage}[b]{\linewidth}\raggedright
Contrast
\end{minipage} & \begin{minipage}[b]{\linewidth}\raggedleft
First method wins
\end{minipage} & \begin{minipage}[b]{\linewidth}\raggedleft
Signed-rank \(p_{Holm}\)
\end{minipage} & \begin{minipage}[b]{\linewidth}\raggedleft
Sign-test \(p_{Holm}\)
\end{minipage} & \begin{minipage}[b]{\linewidth}\centering
Robust
\end{minipage} \\
\midrule\noalign{}
\endhead
\bottomrule\noalign{}
\endlastfoot
Ridge vs XGBoost & 100/160 & 0.0078 & 0.0254 & Yes \\
Ridge vs CatBoost & 82/160 & 1.0000 & 0.9423 & No \\
Ridge vs RealMLP & 70/160 & 0.0211 & 0.7969 & No \\
Ridge vs TabSwift & 52/160 & \(<10^{-4}\) & 0.0002 & Yes \\
Ridge vs TabICL & 47/160 & \(<10^{-4}\) & \(<10^{-4}\) & Yes \\
Ridge vs TabDPT & 37/160 & \(<10^{-4}\) & \(<10^{-4}\) & Yes \\
Ridge vs TabPFN v2 & 55/160 & \(<10^{-4}\) & 0.0018 & Yes \\
Ridge vs TabFM & 39/160 & \(<10^{-4}\) & \(<10^{-4}\) & Yes \\
Ridge vs TabPFN 3.5 & 35/160 & \(<10^{-4}\) & \(<10^{-4}\) & Yes \\
XGBoost vs CatBoost & 37/160 & \(<10^{-4}\) & \(<10^{-4}\) & Yes \\
XGBoost vs RealMLP & 28/160 & \(<10^{-4}\) & \(<10^{-4}\) & Yes \\
XGBoost vs TabSwift & 25/160 & \(<10^{-4}\) & \(<10^{-4}\) & Yes \\
XGBoost vs TabICL & 26/160 & \(<10^{-4}\) & \(<10^{-4}\) & Yes \\
XGBoost vs TabDPT & 16/160 & \(<10^{-4}\) & \(<10^{-4}\) & Yes \\
XGBoost vs TabPFN v2 & 20/160 & \(<10^{-4}\) & \(<10^{-4}\) & Yes \\
XGBoost vs TabFM & 19/160 & \(<10^{-4}\) & \(<10^{-4}\) & Yes \\
XGBoost vs TabPFN 3.5 & 16/160 & \(<10^{-4}\) & \(<10^{-4}\) & Yes \\
CatBoost vs RealMLP & 51/160 & \(<10^{-4}\) & 0.0001 & Yes \\
CatBoost vs TabSwift & 41/160 & \(<10^{-4}\) & \(<10^{-4}\) & Yes \\
CatBoost vs TabICL & 37/160 & \(<10^{-4}\) & \(<10^{-4}\) & Yes \\
CatBoost vs TabDPT & 23/160 & \(<10^{-4}\) & \(<10^{-4}\) & Yes \\
CatBoost vs TabPFN v2 & 40/160 & \(<10^{-4}\) & \(<10^{-4}\) & Yes \\
CatBoost vs TabFM & 34/160 & \(<10^{-4}\) & \(<10^{-4}\) & Yes \\
CatBoost vs TabPFN 3.5 & 28/160 & \(<10^{-4}\) & \(<10^{-4}\) & Yes \\
RealMLP vs TabSwift & 67/160 & 0.1739 & 0.4299 & No \\
RealMLP vs TabICL & 56/160 & 0.0009 & 0.0031 & Yes \\
RealMLP vs TabDPT & 39/160 & \(<10^{-4}\) & \(<10^{-4}\) & Yes \\
RealMLP vs TabPFN v2 & 58/160 & 0.0105 & 0.0101 & Yes \\
RealMLP vs TabFM & 47/160 & \(<10^{-4}\) & \(<10^{-4}\) & Yes \\
RealMLP vs TabPFN 3.5 & 40/160 & \(<10^{-4}\) & \(<10^{-4}\) & Yes \\
TabSwift vs TabICL & 72/160 & 0.2675 & 0.9423 & No \\
TabSwift vs TabDPT & 53/160 & \(<10^{-4}\) & 0.0005 & Yes \\
TabSwift vs TabPFN v2 & 72/160 & 1.0000 & 0.9423 & No \\
TabSwift vs TabFM & 60/160 & 0.0002 & 0.0254 & Yes \\
TabSwift vs TabPFN 3.5 & 43/160 & \(<10^{-4}\) & \(<10^{-4}\) & Yes \\
TabICL vs TabDPT & 66/160 & 0.1456 & 0.3247 & No \\
TabICL vs TabPFN v2 & 91/160 & 0.1907 & 0.6760 & No \\
TabICL vs TabFM & 68/160 & 0.1700 & 0.5495 & No \\
TabICL vs TabPFN 3.5 & 50/160 & 0.0004 & 0.0001 & Yes \\
TabDPT vs TabPFN v2 & 101/160 & \(<10^{-4}\) & 0.0157 & Yes \\
TabDPT vs TabFM & 90/160 & 1.0000 & 0.7969 & No \\
TabDPT vs TabPFN 3.5 & 64/160 & 0.2675 & 0.1540 & No \\
TabPFN v2 vs TabFM & 58/160 & 0.0039 & 0.0101 & Yes \\
TabPFN v2 vs TabPFN 3.5 & 55/160 & \(<10^{-4}\) & 0.0018 & Yes \\
TabFM vs TabPFN 3.5 & 72/160 & 0.4555 & 0.9423 & No \\
\end{longtable}
}

Task identity accounted for \(\eta^2=0.5421\) of variation in the six-FM mean advantage over ridge (permutation \(p=0.0008\)); feature-set identity accounted for 0.0823 and did not exceed its within-task restricted null (\(p=0.1608\)). These 5,000-permutation analyses are descriptive; cells within a task are dependent. Task-averaged advantage showed no clear correlation with cohort size (Spearman \(\rho=0.073\), \(p=0.609\)), test size (\(\rho=0.076\), \(p=0.591\)), or the demographic baseline (\(\rho=0.120\), \(p=0.396\)).

The cell-median advantage over ridge was 0.0043 \(R^2\); its 95\% interval was 0.0029--0.0062 under cell resampling and 0.0026--0.0072 when whole tasks were resampled. Giving each of the 52 tasks equal weight, the mean advantage was 0.0103 (task-bootstrap 95\% interval, 0.0071--0.0136), positive in 43 tasks. Giving the two TabPFN versions one combined vote yielded a mean advantage of 0.0103 (0.0073--0.0136), positive in 44 tasks. The task-level 45-pair family is retained in the statistical artifact. Tasks still share participants and related outcomes; these checks do not establish independent population replications.

\subsection{Same-signal time-series comparison}\label{same-signal-time-series-comparison}

Panel D compares a frozen zero-shot Chronos-2 representation with engineered features from the same input signal, both read by ridge. It does not compare raw signals and does not evaluate a tabular foundation model on the embeddings.

{\def\LTcaptype{none} % do not increment counter
\begin{longtable}[]{@{}
  >{\raggedright\arraybackslash}p{(\linewidth - 10\tabcolsep) * \real{0.1364}}
  >{\raggedright\arraybackslash}p{(\linewidth - 10\tabcolsep) * \real{0.1364}}
  >{\raggedleft\arraybackslash}p{(\linewidth - 10\tabcolsep) * \real{0.1818}}
  >{\raggedleft\arraybackslash}p{(\linewidth - 10\tabcolsep) * \real{0.1818}}
  >{\raggedleft\arraybackslash}p{(\linewidth - 10\tabcolsep) * \real{0.1818}}
  >{\raggedleft\arraybackslash}p{(\linewidth - 10\tabcolsep) * \real{0.1818}}@{}}
\toprule\noalign{}
\begin{minipage}[b]{\linewidth}\raggedright
Signal
\end{minipage} & \begin{minipage}[b]{\linewidth}\raggedright
Matching unit
\end{minipage} & \begin{minipage}[b]{\linewidth}\raggedleft
Pairs
\end{minipage} & \begin{minipage}[b]{\linewidth}\raggedleft
Chronos-2 better
\end{minipage} & \begin{minipage}[b]{\linewidth}\raggedleft
Signed-rank \(p\)
\end{minipage} & \begin{minipage}[b]{\linewidth}\raggedleft
Sign-test \(p\)
\end{minipage} \\
\midrule\noalign{}
\endhead
\bottomrule\noalign{}
\endlastfoot
CGM & Task and cohort & 31 & 6 & 0.00159 & 0.000878 \\
Sleep & Task and exact participant cohort & 32 & 2 & \(4.66\times10^{-9}\) & \(2.46\times10^{-7}\) \\
ECG & No engineered ECG predictor exists & 0 & -- & -- & -- \\
\end{longtable}
}

The results were insensitive to the most negative pair: after dropping the fatty-liver-index comparison, the signed-rank \(p\) values were 0.00299 for glucose and \(9.31\times10^{-9}\) for sleep. The engineered-feature advantage narrowed with training size (Spearman correlation between the Chronos-2-minus-engineered gap and training size: 0.594 for glucose and 0.458 for sleep). In the larger half of each modality, the signed-rank result was not significant for glucose (16 pairs; \(p=0.597\)) but remained significant for sleep (16 pairs; \(p=0.000214\)). Thus Panel D combines representation differences with the sample efficiency of fitting a ridge head to 770--3,076-dimensional embeddings.

The exploratory binary extension contains 12 matched cells across six tasks and all nine methods, using September baseline runs on the same realized cohorts as the added FMs. The Friedman test gives \(p=1.40\times10^{-5}\); 7 of 36 pairwise contrasts survive both Holm-adjusted tests. This small, dependent set is reported separately from the regression figure. The type-2-diabetes DXA cell has 13 test positives, below the benchmark's 17-per-class evidential threshold: it remains in this exploratory analysis but is refused board publication. The Figure 3 synthesis is post hoc; paired between-method uncertainty within individual cells is not estimated by the across-cell or across-task intervals.

\section{Machine-querying PhenoBench evidence}\label{machine-querying-phenobench-evidence}

This appendix reports an additional evaluation of whether tool-using language models can query a frozen PhenoBench evidence snapshot. Extended Data Figure 1 summarizes the experiment.

\subsection{Results}\label{results-1}

We tested whether tool-using language models could retrieve and compare PhenoBench evidence while preserving the conditions that make each comparison valid. The questions tested five operations: retrieving a displayed result; selecting a method while holding the task and information source fixed; selecting an information source while holding the task and method fixed; combining a Task Card constraint with leaderboard evidence; and deciding whether two rows were comparable under the benchmark contract (Extended Data Figure 1 and Evaluation design below). Answers had to include the supporting rows, allowing conclusion and evidence fidelity to be assessed separately. Outputs were graded deterministically against 277 reviewed reference answers and independently recomputed from raw model transcripts.

High exact fidelity was feasible but not general across the tested models. Claude Sonnet 4.5 achieved mean accuracies of 94.9\% for descriptive lookup, 93.1\% for method selection with complete candidate evidence, 88.9\% across the nine eligible information-source cases, and 91.7\% for context-integrated selection. Other models were uneven: for example, Claude Opus 4.8 achieved 96.4\% lookup accuracy but 51.1\% method-selection accuracy, whereas Claude Haiku 4.5 achieved 58.7\% and 83.9\%, respectively. These differences show that reading a result, selecting within a controlled comparison and preserving its evidence are distinct capabilities.

Task Card access increased context-question accuracy for six of seven models by 22.9--68.8 percentage points relative to the leaderboard-only control. This effect is deliberately narrow: the questions required one reviewed card fact that was absent from the leaderboard. Pixtral was the exception, falling from 27.1\% without cards to 0\% with cards. Transcript inspection showed a model--tool interoperability failure: after reading the card, 46 of 48 runs serialized the next leaderboard request as answer text rather than issuing a tool call. Explicit comparison rules showed a related distinction between conclusions and evidence. Six models returned the correct supported/unsupported verdict on every rule item, but their exact verdict-plus-reason-plus-evidence fidelity ranged from 25.0\% to 100\%; Pixtral achieved 77.1\% verdict fidelity and 24.0\% full fidelity.

These results establish that querying benchmark evidence is itself measurable: some tested models achieved high exact fidelity, whereas others failed selectively at comparison, evidence preservation or tool use. They do not establish open-ended scientific reasoning, clinical decision quality or general model superiority.

\begingroup
\renewcommand{\figurename}{Extended Data Figure}
\setcounter{figure}{0}
\begin{figure*}[t]
\centering
\includegraphics[width=\textwidth]{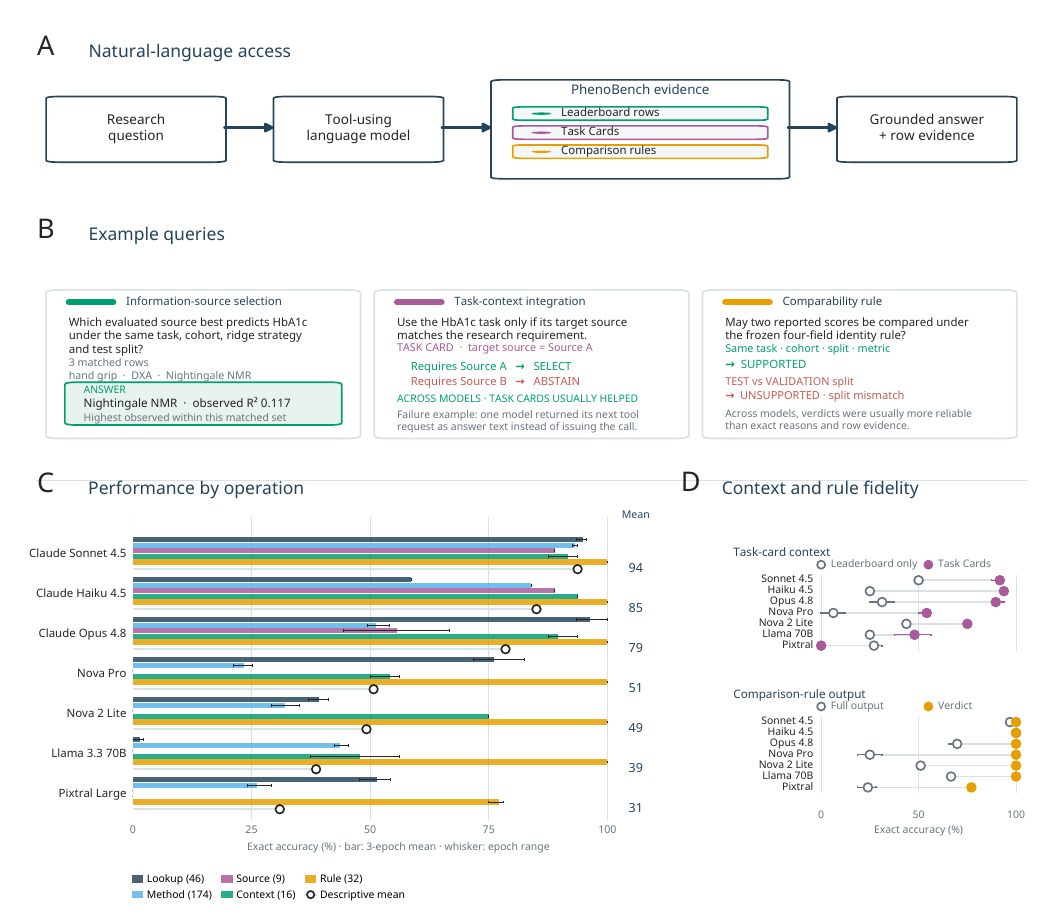}
\caption{\textbf{Tool-using language models query PhenoBench evidence with model-dependent reliability.} \textbf{A,} A tool-using language model receives a natural-language research question, inspects frozen leaderboard rows, Task Cards and comparison rules, and returns a grounded answer with row evidence. Outputs were checked against 277 reviewed reference answers by deterministic grading and independent replay. \textbf{B,} Shortened examples show controlled information-source selection, Task Card-dependent selection or abstention, and a valid/invalid comparison-rule pair. \textbf{C,} Grouped bars show each model's mean exact accuracy across three epochs for descriptive lookup ($n=46$), controlled method selection ($n=174$), controlled information-source selection ($n=9$), Task Card-integrated selection ($n=16$) and comparison-rule full-output fidelity ($n=32$); whiskers show the epoch range. Open circles and printed values give the equal-weight arithmetic mean of the five operation means. This mean was added post hoc for visual compression, is descriptive only and is not a prespecified endpoint or general model ranking. \textbf{D,} Task Card access is compared with leaderboard-only access for the context-dependent questions, and exact comparison verdicts are compared with full verdict-plus-reason-plus-evidence fidelity. Task Cards improved context accuracy for six models but exposed a Pixtral tool-loop failure; comparison verdicts were generally more reliable than full outputs. Reference answers and source facts were reviewed before the run, and a second implementation parsed raw completions and independently recomputed all outcomes. Results are limited to the tested models, prompts, tools and frozen snapshot.}
\label{fig:machine-operability-extended}
\end{figure*}
\endgroup

\subsection{Evaluation design}\label{evaluation-design}

We tested whether tool-using language models could answer natural-language questions from a frozen, version-pinned PhenoBench evidence snapshot. The tools returned leaderboard rows, Task Cards, or the two rows in a specified comparison case; the models had no other route to benchmark data. A manually reviewed bank contained 277 questions: 46 descriptive leaderboard lookups, 174 method comparisons that fixed the task, information source and evaluation identity, 9 information-source comparisons that fixed the task, method and evaluation identity, 16 questions that required combining a Task Card fact with leaderboard evidence, and 16 minimal pairs (32 questions) testing an explicit four-field comparability rule. The nine information-source questions exhaust the eligible controlled comparisons in the frozen snapshot and are therefore interpreted cautiously.

We implemented the evaluation in Inspect AI v0.3.258 {[}21{]}, which recorded the prompts, tool calls, model outputs and scoring results for every sample. We evaluated seven Bedrock-hosted model versions: Claude Sonnet 4.5, Claude Haiku 4.5, Claude Opus 4.8, Nova Pro, Nova 2 Lite, Llama 3.3 70B and Pixtral Large 25.02. Each model answered every question in three epochs. Temperature was zero where the provider exposed that setting; the Opus endpoint did not. Prompts, tool definitions, maximum output length, message limit and deterministic grader were otherwise fixed. Sustained quota failures prevented completion through the prespecified US Pixtral inference profile, so before starting a complete Pixtral run we prospectively amended only its AWS routing profile from US to EU; the underlying \texttt{pixtral-large-2502-v1:0} model version and all experimental inputs were unchanged, and incomplete US logs were excluded.

Primary correctness was construct-specific. Descriptive lookup required the exact row, metric name and displayed value. Method and information-source comparisons additionally required the complete candidate-row evidence from the controlled set. Context questions required the correct Task Card fact and, when the requirement was satisfied, the exact leaderboard selection; when it was not satisfied, the model had to return no selection. Comparability questions were scored primarily on the supported/unsupported verdict, with exact reason codes and row evidence reported separately. The context questions were repeated in a leaderboard-only control with the Task Card tool removed. All 277 reference answers were audited before the confirmatory run. A second verifier parsed raw model completions independently of the runtime grader, recomputed every primary outcome, checked exact question coverage and frozen-file hashes, and priced recorded token use.

We report each model and construct separately as the arithmetic mean and range of its three epoch accuracies. Repeated epochs over the same question bank were not pooled as independent observations, and no item-level confidence interval was computed from them. No cross-construct endpoint was prespecified or used for inference. For visual compression only, Extended Data Figure 1C adds a post hoc arithmetic mean that assigns each of the five displayed operations equal weight; it is labeled descriptive, and every component remains visible.

\subsection{Limitations}\label{limitations-1}

The machine-querying experiment measures exact adherence to a frozen benchmark interface, not open-ended scientific reasoning. Reference answers were reviewed against one PhenoBench snapshot and explicit comparison rules, and results may depend on the prompts, tool schema, evaluation harness and provider implementation. Coverage was uneven: information-source selection comprised only nine eligible cases, and the Task Card analysis used eight reviewed facts encoded as matched satisfied and unsatisfied questions. Because these questions were designed to require information absent from the leaderboard, the ablation tests whether models can use necessary Task Card context, not whether Task Cards generally improve research quality. Repeated epochs reused the same items and are not independent samples. Model--tool failures, including Pixtral's serialized tool calls, therefore characterize the tested model--harness combination rather than the model alone. High interface accuracy does not establish biomedical reasoning quality, clinical utility, PhenoBench-LLM performance, autonomous research or general model superiority.

\clearpage

\section{PhenoBench-LLM supplementary evidence}\label{phenobench-llm-supplementary-evidence}

\subsection{Extended results: PhenoBench-LLM}\label{extended-results-phenobench-llm}

We next used cohort data to evaluate language models under controlled, clinically grounded conditions. Fourteen frontier language models from eight developers, spanning flagship, mid-size and small tiers, were asked the same HPP questions from the same per-participant evidence packet (demographics, CGM summaries, Nightingale NMR metabolomics, and optionally the Task Card) under a frozen prompt, and their answers entered PhenoBench as Imported Prediction Artifacts scored under 40 task formulations in four categories: recovering a hidden measured phenotype (25 Tasks, \(R^2\)), classifying a curated status (5 Tasks, AUROC), forecasting a measurement at the two-year follow-up visit from the baseline packet (4 Tasks, \(R^2\)), and ordering groups of four participants (6 Tasks, pairwise accuracy), on 150--480 validation participants per Task (Figure~\ref{fig:phenobench-llm}, Supplementary Figure~\ref{fig:phenobench-llm-contract} and Methods). Every cell is one Eval Run whose row-level output was verified against the artifact, and every number was reproduced by an independent implementation.

Cohort Tasks revealed distinct capability profiles (Figure~\ref{fig:phenobench-llm}A). By pairwise win rate over all available head-to-heads, Gemini 3.7 Flash won 79\% (95\% CI 71--86), Claude Sonnet 5 70\% (61--77) and Claude Opus 4.8 68\% (59--77), while Llama 3.3 70B and Pixtral Large won 32\% and GPT-5.4 nano 30\% (21--39). The ordering was similar on the eleven Tasks run by every model (Gemini 3.7 Flash 83\%, Opus 4.8 73\%, Sonnet 5 73\%; GPT-5.4 nano 21\%), but category profiles were not interchangeable. Opus 4.8 led classification head-to-heads (82\%), Gemini 3.7 Flash led phenotype recovery (76\%), follow-up forecasting (86\%) and ordering (91\%), and GPT-5.4 nano rose from 23--33\% in phenotype recovery and classification to 80\% in follow-up forecasting. This longitudinal win rate compares language models with one another on raw follow-up \(R^2\) and does not measure improvement over carry-forward. Category estimates use available same-Task comparisons and therefore retain unequal coverage.

Cost alone did not determine performance (Figure~\ref{fig:phenobench-llm}B). On the eleven shared Tasks, the efficiency frontier comprised Llama 4 Maverick (40\% win rate; \$0.00042 per full-packet call), GPT-5.6 (62\%; \$0.00065) and Gemini 3.7 Flash (83\%; \$0.0030). GPT-5.4 nano was dominated by Llama 4 Maverick on both cost and win rate (21\%; \$0.00063 per call) and therefore was not on the frontier. Gemini 3.7 Flash outperformed every more expensive model, including Opus 4.8 at \$0.016 per call. Full cost, reliability and all-Task estimates are shown in Supplementary Figure~\ref{fig:phenobench-llm-model-detail}.

Fitted probes provided a reference for interpreting these capability profiles. On phenotype recovery the median full-packet \(R^2\) of the best model was 0.04 and of the weakest \(-0.34\); models exceeded the demographic floor on at most 5 of 25 Tasks (Opus 4.8) and exceeded the ridge probe over the same fields in 8 of 304 paired model--Task comparisons. In classification the best models matched the demographic classifier (median AUROC 0.75 versus a floor of 0.77) and exceeded the logistic comparator in 10 of 66 cells, most of them on migraine, where that comparator itself fell below the floor. In ordering no model's pairwise accuracy exceeded the ordering induced by the ridge probe on five of six Tasks. The language models received each participant's baseline measurement for follow-up Tasks, whereas the demographic floor did not. Compared with simply carrying the baseline value forward, models added \(R^2\) of \(+0.02\) to \(+0.10\) on systolic blood pressure but \(-0.06\) to \(+0.06\) on the apnea--hypopnea index, and the ridge probe with the same fields added more. Absolute per-Task estimates and probe references are shown in Supplementary Figures~\ref{fig:phenobench-llm-model-detail} and~\ref{fig:phenobench-llm-task-deltas}.

The same controlled Tasks exposed failure modes that aggregate rankings conceal. Across the 11-Task map, the median Task-Card change in Spearman correlation ranged from \(-0.017\) to \(+0.026\) across four models, while the median change in \(R^2\) ranged from 0.000 to \(+0.019\). These aggregate changes coexisted with task-specific units and target-semantics failures: on retinal artery width every model answered on a micrometre-like scale against a target in AutoMorph pixels (bias \(-18{,}500\) pixels, spread near zero), and on liver attenuation most models placed estimates on the wrong scale. The reference-containing v1 Task Card repaired the liver-attenuation scale; the sanitized v2 card used in the suite did not reproduce that calibration, and retinal artery width remained off-scale under both card versions. Parse failures were rare (below 0.2\% for every retained model) and Pixtral Large lost 10.7\% of calls to transport errors, which were excluded rather than imputed (Supplementary Figure~\ref{fig:phenobench-llm-model-detail}).

This evaluation used one run per cell, 150--480 validation participants per Task, a fixed packet of summary features and a single prompt family. The contribution is a controlled cohort-data evaluation of model capability profiles, efficiency tradeoffs and clinically relevant failure modes. Fitted-probe comparisons calibrate how much predictive information the language models extract from the same evidence.

\begingroup
\renewcommand{\figurename}{Supplementary Figure}
\setcounter{figure}{1}
\begin{figure*}[t]
\centering
\includegraphics[width=\textwidth]{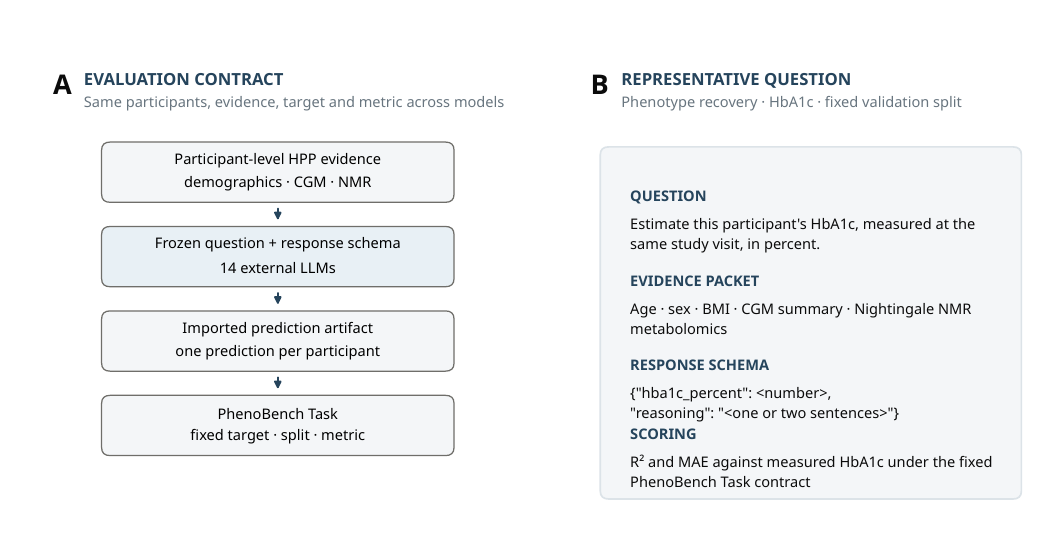}
\caption{\textbf{PhenoBench-LLM evaluation contract and representative question.} \textbf{A,} Every model receives the same participant-level HPP evidence under a frozen question and response schema. Predictions are stored as Imported Prediction Artifacts and scored against the fixed target, validation split and metric owned by the PhenoBench Task. \textbf{B,} Representative HbA1c phenotype-recovery question and response schema. The evidence fields and wording are the task template; no participant row or model response is reproduced.}
\label{fig:phenobench-llm-contract}
\end{figure*}

\begin{figure*}[t]
\centering
\includegraphics[width=\textwidth,height=0.82\textheight,keepaspectratio]{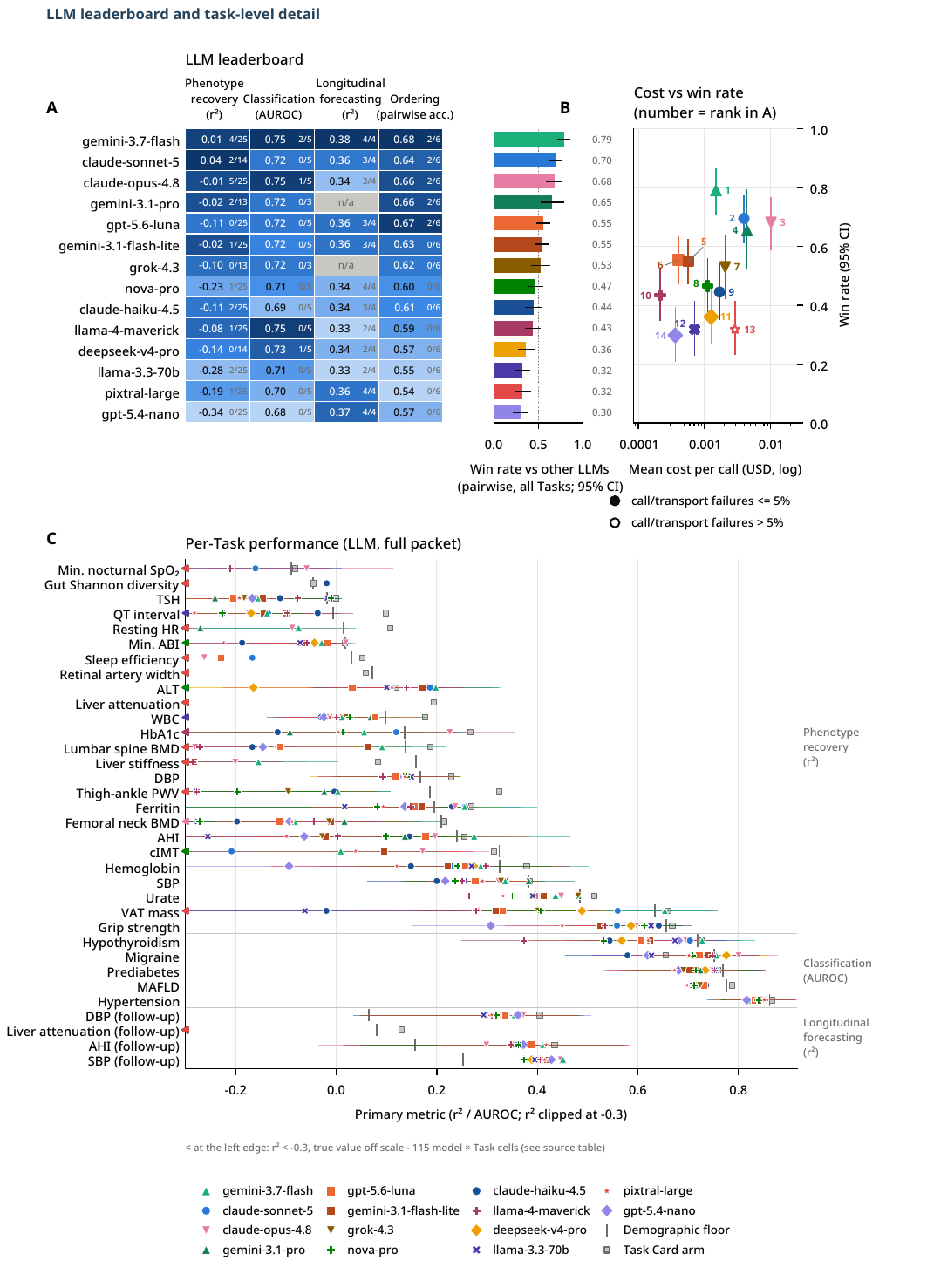}
\caption{\textbf{PhenoBench-LLM leaderboard and Task-level detail.} \textbf{A,} Category leaderboard and all-Task pairwise win rate; cells report median full-packet performance and the fraction of Tasks beating the demographic floor, except for follow-up forecasting, where the fraction is relative to baseline carry-forward. \textbf{B,} Mean cost per call against all-Task win rate; numbers index the model rows in Panel A and intervals are Task-bootstrap 95\% confidence intervals. \textbf{C,} Absolute full-packet performance for every model and point-estimation Task, with participant-bootstrap intervals, demographic floors and fitted-probe references. Off-scale $R^2$ values are retained in the source table and marked at the axis edge.}
\label{fig:phenobench-llm-model-detail}
\end{figure*}

\begin{figure*}[t]
\centering
\includegraphics[width=\textwidth,height=0.82\textheight,keepaspectratio]{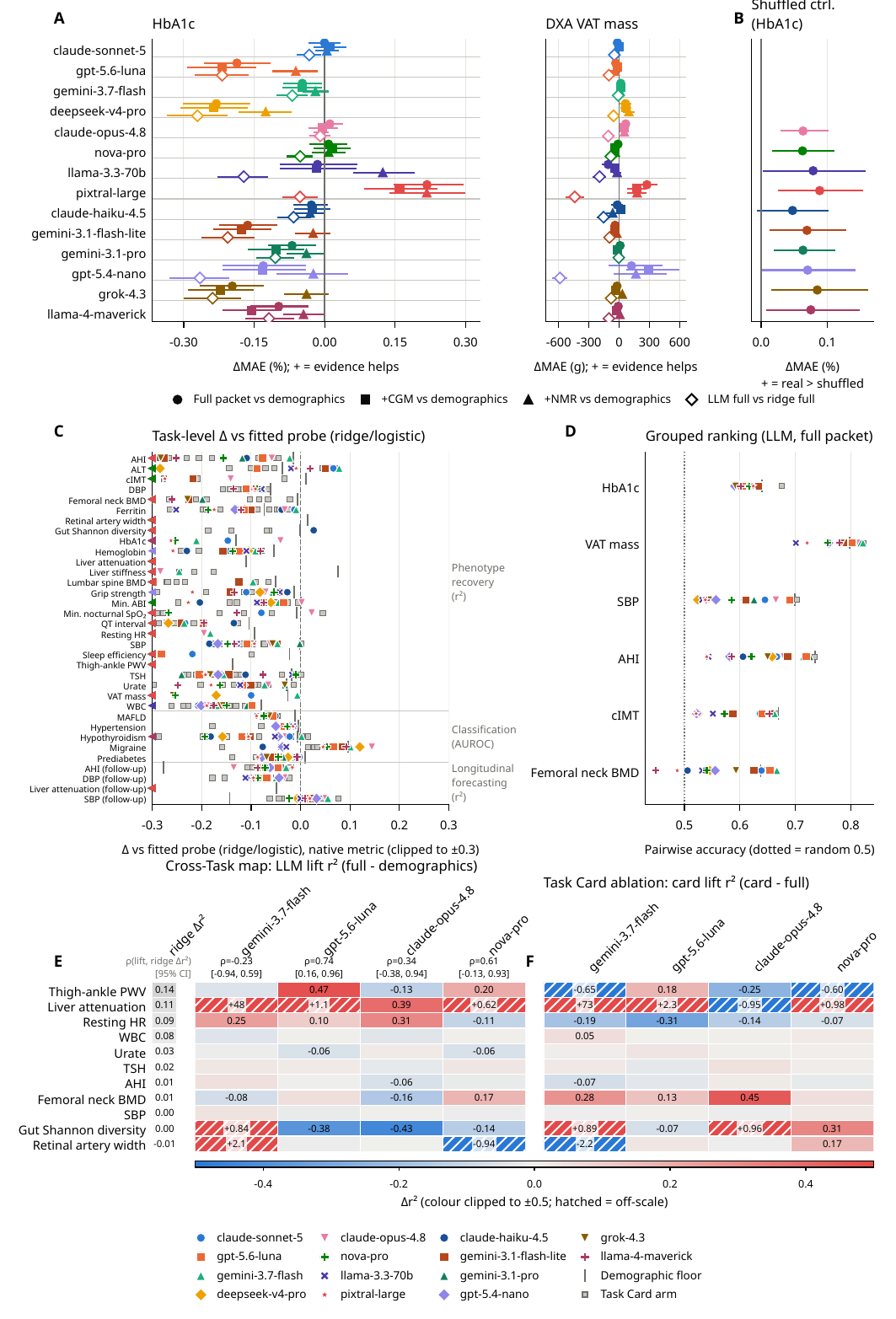}
\caption{\textbf{PhenoBench-LLM paired and mechanistic analyses.} \textbf{A,} Paired MAE effects of evidence arms and the fitted probe for HbA1c and VAT mass. \textbf{B,} HbA1c shuffled-packet control. \textbf{C,} Per-Task full-packet difference from the fitted probe. \textbf{D,} Grouped-ranking pairwise accuracy with random, demographic-order and ridge-order references. \textbf{E,} Cross-Task map of full-packet $R^2$ change from demographics, ordered by the fitted-probe $R^2$ change. \textbf{F,} sanitized Task Card $R^2$ change from the full packet. Off-scale heatmap values are printed and hatched. All comparisons use the validation split and retain null and negative estimates.}
\label{fig:phenobench-llm-mechanisms}
\end{figure*}

\begin{figure*}[t]
\centering
\includegraphics[width=\textwidth]{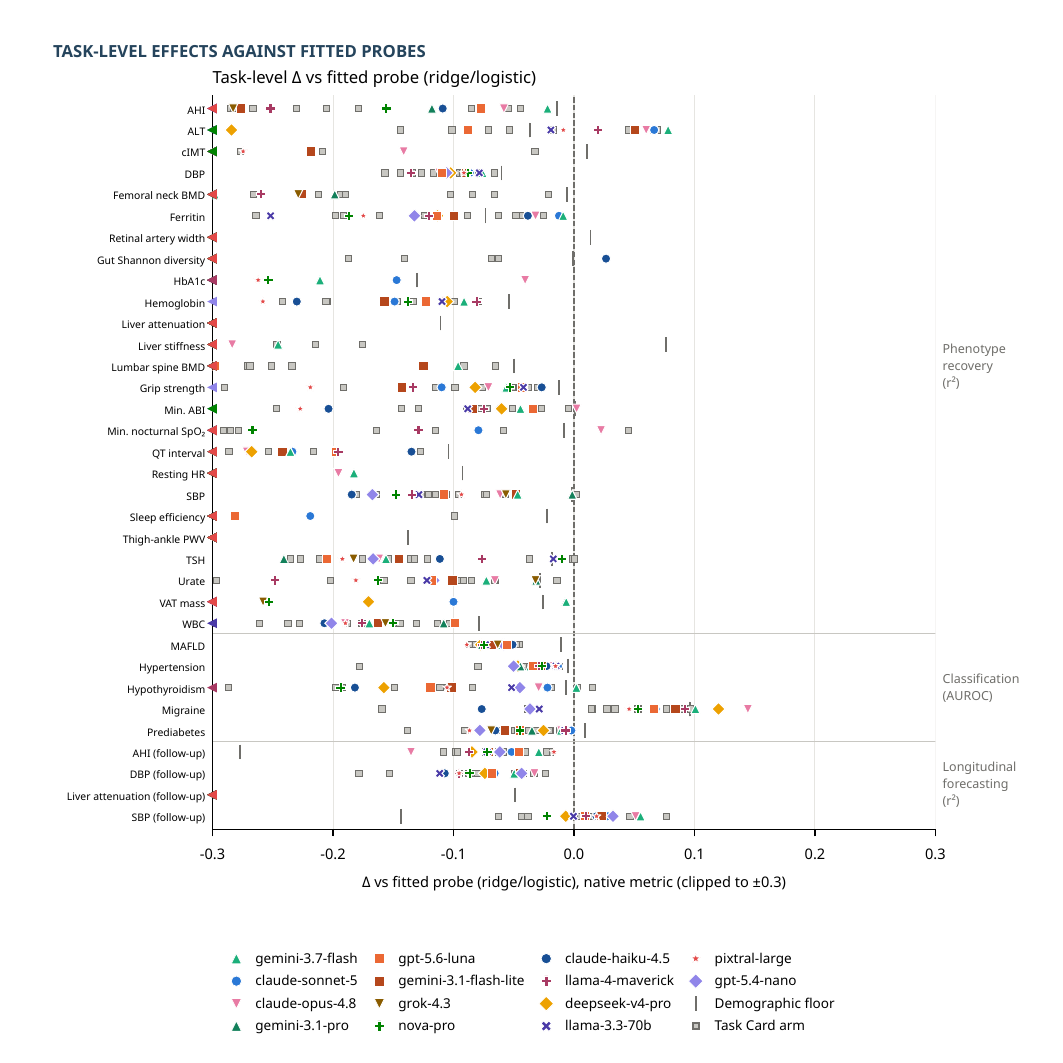}
\caption{\textbf{Task-level PhenoBench-LLM effects against fitted probes.} Each marker is one model's full-packet primary-metric difference from the ridge or logistic probe over the same fields; zero denotes the fitted probe. Grey ticks show the demographic-floor difference from the probe and grey squares show the Task Card arm where available. Values outside $\pm0.3$ are marked at the edge and retained exactly in the source table. Tasks are grouped by phenotype recovery, classification and longitudinal forecasting; grouped ranking is shown in Supplementary Figure~\ref{fig:phenobench-llm-mechanisms}.}
\label{fig:phenobench-llm-task-deltas}
\end{figure*}
\clearpage
\endgroup

\section{Extended methods}\label{extended-methods}

\subsection{Cohort and measurements}\label{cohort-and-measurements}

The Human Phenotype Project (HPP) is a prospective longitudinal cohort with broad clinical, molecular, imaging, and wearable phenotyping {[}19{]}. Measurements include medical history and lifestyle, anthropometrics, blood assays, continuous glucose monitoring (CGM), overnight sleep physiology, imaging, and multi-omic assays. PhenoBench analyses use task-specific eligible subsets rather than one common cohort: each Task defines its target, timepoint, required measurements, quality-control rules, exclusions, and evaluation grain.

Unless a Task specifies a documented alternative, participants are assigned by a stable participant-identifier hash to canonical 70/15/15 train, validation, and test splits (\texttt{participant\_id\_split\_v1:0.7,0.15,0.15}). Repeated observations are grouped by participant so that one person does not cross splits. Imputation and fitted preprocessing use training data only; model and hyperparameter development use training and validation data. The held-out test split is reserved for reviewed final results. Event- or artifact-based Tasks record any different grouping or split policy in their executable contract.

HPP participants provided informed consent, and identifying details were removed before computational analysis. The study was conducted according to the Declaration of Helsinki and approved by the Weizmann Institute of Science Institutional Review Board (approval 2392-4) {[}19{]}. PhenoBench reports aggregate evaluation results and does not release individual-level participant data.

\subsection{Benchmark tracks and model comparisons}\label{benchmark-tracks-and-model-comparisons}

We defined a model-comparison cell by task, cohort, predictor, and evaluation split. Within a cell, the target, participants, feature set, and split were fixed while the fitting method varied. The regression capacity analysis used complete-case cells in which all ten evaluated methods were available: ridge, XGBoost, CatBoost, RealMLP, TabSwift, TabICL, TabDPT, TabPFN v2, Google TabFM, and TabPFN 3.5. This yielded 160 test-set comparison cells spanning 52 tasks. Execution revisions are preserved per run. The added models were audited against the same configuration, feature specification, realized cohorts, and train/validation/test participant hashes as the matched references. The release review explicitly binds the distinct source revisions and common scientific runtime; no execution identity is rewritten.

Ridge was fitted conventionally on the task training data. XGBoost 3.2.0 and CatBoost 1.2.10 each used five-fold cross-validation over the same fixed 12-candidate hyperparameter grid within the training split; the selected configuration was then scored once on the held-out test split. For each library, the 12 candidates were the Cartesian product of 100, 300, or 600 boosting iterations, learning rates of 0.03 or 0.1, and maximum depths of 3 or 6. Regularization was fixed at 1.0 (XGBoost \texttt{reg\_lambda}; CatBoost \texttt{l2\_leaf\_reg}), and the model random seed was 42. RealMLP {[}26{]} was likewise trained from scratch (\texttt{pytabkit} 1.7.3); it is included as a strong neural tabular baseline, not labeled as a pretrained foundation model. TabSwift {[}20{]}, TabICL {[}6{]}, TabDPT {[}27{]}, TabPFN v2 {[}32{]}, Google TabFM {[}33{]}, and TabPFN 3.5 {[}34{]} used frozen pretrained checkpoints under their pinned benchmark implementations. All methods received the same columns and participants within a cell; model-specific preprocessing and ensembling followed the persisted run configuration.

For each method, we calculated \(\Delta R^2\) as its full-model test \(R^2\) minus the common demographics-only ridge baseline from the matched ridge run. Demographics comprised age, sex, and BMI, with age omitted when chronological age was the target. Using one baseline made within-cell ranks identical to ranks based on raw \(R^2\). For non-ridge methods, this common reference means that \(\Delta R^2\) combines differences in fitting demographics with differences in using the added predictors. Cells missing any of the ten methods were excluded from the complete-case capacity analysis.

The time-series comparison used frozen, zero-shot Chronos-2 embeddings of CGM, electrocardiography (ECG), and overnight sleep recordings with a ridge head. Each embedding route was compared only with features engineered from the same signal. The original candidate grid added 52 engineered-feature configurations derived from corresponding Chronos-2 configurations by changing only the predictor: 31 for CGM and 21 for WatchPAT. Task arguments, split, strategy, and penalty grid were retained. The 31 glucose pairs selected 21 generated and 10 pre-existing engineered runs and were matched on task and exact participant cohort. The earlier task-only WatchPAT pairs were replaced by 32 post hoc exact-intersection reruns. For WatchPAT, the frozen embedding export supported 5,727 participants and the curated sleep-summary panel supported 9,653; their intersection contained 5,574 participants, or 97.3\% of embedding support. A fresh warehouse read traced the 153 embedding-only participants to six absent source records, 92 participants without a successful sleep pipeline, and 55 with incomplete summaries after quality control, with zero unexplained exclusions. Each task's eligible embedding cohort was intersected with this shared support and persisted as a content-addressed cohort artifact used by both arms. Paired arms retained the canonical split, ridge strategy, and penalty grid and differed only in predictor. The WatchPAT rerun was excluded from frozen leaderboards. ECG contributed no pair because no engineered ECG feature set exists in the corpus. This analysis therefore compared representation routes under a linear head; it did not test whether a tabular foundation model could use the embeddings differently.

\subsection{Statistical analysis}\label{statistical-analysis}

For the ten-method analysis, we ranked methods within each comparison cell (rank 1 denotes the highest \(\Delta R^2\)). Rank point estimates describe the finite benchmark; their percentile 95\% intervals use 5,000 bootstrap draws of whole tasks, retaining all cells belonging to each sampled task (seed 20260828). The cell-median six-FM advantage over ridge used the same whole-task resampling. The primary cross-task summary first averaged each model's scores within task so each of the 52 tasks received equal weight; its interval resampled those task means. Equal-family sensitivity averaged the two TabPFN versions before giving each of five FM families equal weight. Task-level contrasts use one mean per task. Cell-level tests were descriptive: Friedman tests, the ordering-free within-cell mean contrast, Page's grouped-order sensitivity over all 720 FM permutations, and all 45 paired method contrasts are retained as finite-benchmark summaries. Pairwise families use Wilcoxon signed-rank and exact binomial sign tests, corrected separately by Holm; exact ties do not enter the sign-test denominator. Signed-rank inference assumes symmetric paired differences, whereas sign tests use only their directions. Panel B shows descriptive Tukey box plots with 1.5-IQR whiskers and omitted outliers.

We quantified task- and feature-set-associated variation using \(\eta^2\) and 5,000 permutations. Task labels were permuted freely; feature-set labels were permuted within task because feature set was nearly nested in task. The 160 cells describe registered task--track combinations and should not be read as independent clinical questions. For correlates of model advantage, we first aggregated to the 52 task-level units. For each same-signal Chronos-2 comparison, we applied a paired Wilcoxon signed-rank test and an exact binomial sign test on units matched by task and exact participant cohort.

\subsection{Clinically grounded evaluation tasks}\label{clinically-grounded-evaluation-tasks}

We began by asking which clinical and scientific questions could be answered using the HPP and which measurements could be used to answer them. The first source was selected publications based on the HPP cohort: these studies provided questions that had already been answered, or that investigators had attempted to answer, using the same participants and measurements. A clinician-authored inventory of surrogate targets provided an additional starting point, and later benchmark work broadened the collection beyond questions represented in the initial papers. These additions complemented the publication-led ingestion process rather than defining a separate primary route.

We first piloted this review on selected HPP publications and then applied it to a curated corpus of 61 materialized HPP paper objects. An agent-assisted pass used paper metadata, structured summaries, prior replication records, and targeted full-text checks. Six paper objects lacked enough body text or summary material for rigorous interpretation and were marked as evidence gaps rather than treated as studies with no usable task. The screen produced 31 candidate asset families. These counts describe the ingestion audit, not the provenance of every task in the final 90-task collection. Detailed audit outcomes are reported in the Appendix.

The paper review used structured extraction to turn prose into candidate task definitions. For each relevant result, we recorded a short task name, the clinical or scientific question, target, cohort and timepoint, information source, model or analytic strategy, metric, reported result, and split protocol. This step was closer to identifying and linking the parts of a question than to summarizing the paper as a whole. We did not translate each paper, or each extracted result, into a new Task.

We next assembled the clinical context needed to interpret each candidate. This included the source publication, targeted literature searches supported by Paperclip, and relevant OpenEvidence query responses that were manually examined by a clinician, the same physician who reviewed the Task Cards. We used this material to compose Task Cards describing why each target matters, how it is measured in HPP, meaningful baselines and comparators, leakage and confounding risks, measurement noise, suitable metrics, relevant evidence, and the claims a result could and could not support. In the converged process, each human-readable card was reviewed before the executable task was finalized.

We then mapped the extracted question onto the PhenoBench components: Task, Predictor, Strategy, configuration, metric, artifact, and evaluation record. A distinct target, target construction, eligible population, or scoring policy could justify a new Task. A different information source, representation, model, or visit pairing normally became a new Predictor, Strategy, or configuration for an existing Task.

Candidate validity and implementation readiness were assessed separately. Validity asked whether the target and proposed comparison formed a coherent scientific or clinical question, including whether the target was measured with the required units and timing, whether the eligible cohort could support the claim, and whether leakage or circularity made the comparison uninterpretable. Readiness asked whether the necessary data route, artifact, target shape, split policy, metric, and provenance already existed. A valid candidate was therefore retained as future capability or artifact work when it could not yet be executed safely; missing software support was not itself a reason to reject the question.

For accepted targets, the executable Task specifies target construction and units, the unit of evaluation, exclusions and quality control, timing, participant grouping and the split policy. A Benchmark Track records the allowed inputs, which split is scored, and the primary metric. We fixed these settings and the comparator before interpreting results. A Task can have multiple Tracks so different measurements can be evaluated for the same question.

After implementation, we reconciled the Task Card against the executable Task, configuration, HPP dataset documentation, and original literature. Agent assistance and literature tools accelerated extraction and evidence retrieval, but did not decide whether a question was clinically meaningful or whether a comparison supported a claim. The Task Cards document clinical rationale, measurement definition and units, timing, suitable comparators, leakage risks, and claim boundaries. We do not report independent specialist ratings, inter-rater agreement, or a complete per-card revision/rejection log. Manual clinical examination of the supporting material was distinct from formal domain-expert adjudication of every card; unresolved measurement-sensitive decisions remained explicit, and expert-review status required review by the relevant domain expert.

An implemented candidate was considered reviewable only after synthetic contract checks and a real-data run. We inspected cohort and split counts, target and predictor support, the persisted run manifest, comparison with the declared baseline, and any contamination or provenance warnings. Paper-derived rows were labeled as paper-faithful, paper-inspired analogues, weak, or non-comparable according to whether the cohort, target, features, preprocessing, split, and metric supported the comparison. A runnable configuration alone did not establish that a published result had been reproduced.

\subsection{Model onboarding and model cards}\label{model-onboarding-and-model-cards}

Task curation defines the question; model onboarding defines the system used to answer it. A Model Card documents a model family: an encoder together with the inputs it consumes. A version is a retrain within that family; a model we never retrain carries no version. We wrote cards for contributed models whose output enters an evaluation as a versioned artifact. Predictors defined inside the benchmark, the generic evaluation Strategies, and analysis controls carry no card.

A contributed model may have been pretrained on participants it is later scored on. Each Predictor therefore declares its upstream training exposure from a fixed vocabulary: no upstream model, no HPP training, HPP training restricted to the canonical train split, HPP training on another split, unknown, or undeclared. Where the producer files the list of participants the model trained on, we derive the exposure from that list and intersect it with the cohort the run scored. The ranking gate reads that intersection first and falls back to the declaration only when either list is unknown; a declared HPP-other-split exposure is excluded, an unknown exposure remains explicit, and an undeclared exposure is refused for exportable configurations. Both values stay on the manifest, so a wrong declaration stays visible. Two lists are filed so far; the CGM encoder's spans all three canonical splits (5,198 train, 1,181 validation, 1,178 test), so rows using it are scored partly on participants it trained on.

Contributed models are frozen when evaluated, and any fine-tuning happened upstream under its own model identity. The only components fitted on PhenoBench training data are the probe head and its regularization strength, cross-validated within the training folds. A checksum-bound provenance document names the model, the artifact bytes, the participants, and the information sources the artifact may claim. An evaluation refuses an artifact whose bytes disagree with it. A changed configuration receives a new atomic identifier. We keep the card with the repository that produces the model, and its identifier links a run manifest or published result to it. For future contributions, the card is also the review surface for model source and license, accepted inputs and preprocessing, and the inference environment; these remain narrative rather than code-enforced fields.

Development uses training and validation data. Our release protocol requires a signed review, a single held-out test evaluation, re-derivation of its numbers from the persisted manifests, and a leaderboard commit by a second owner. The card then records that conclusion and the claim boundary: an evaluation measures predictive association, not causal effect or clinical utility, and a representation that recovers what its recording device already measures shows instrument read-back, not phenotype transfer. A card also records what its representation provably cannot encode. For example, Chronos-2 CGM embeddings are invariant to glucose level and variability, so the scored embeddings append the level and scale they discard under their own model identifier; a WatchPAT time series was sampled at 10 Hz, so it keeps pulse rate and loses pulse-wave shape, and a null result does not bound what a shape-preserving encoder could find in the same recording.

External model submissions use an OCI image pinned by an immutable digest, as specified in the public contribution guide (https://github.com/galsapir/phenobench-benchmark/blob/main/CONTRIBUTING.md). Maintainers run accepted images without network access against private task bundles mounted read-only; the image writes predictions to a separate output directory, and those predictions must pass schema and identifier-alignment validation before result curation. Schema-faithful synthetic bundles expose the same input and output contract without releasing participant data.

\subsection{PhenoBench-LLM: external prediction evaluation}\label{phenobench-llm-external-prediction-evaluation}

We evaluated whether frontier language models can use HPP multimodal evidence to predict hidden measured phenotypes beyond their public priors and beyond strong conventional baselines, under fixed PhenoBench contracts. The design was frozen on the canonical validation split before any full-panel inference was scored, and protected test gates were not inspected or consumed. Two existing PhenoBench Tasks were used unchanged: HbA1c (\texttt{blood\_tests.bt\_\_hba1c}, \%) and total visceral adipose tissue mass from DXA (\texttt{dxa.total\_scan\_vat\_mass}, g), both at the baseline visit with the participant as the evaluation unit and \(R^2\) as the primary metric. Both targets are measured at the same visit as CGM and Nightingale NMR metabolomics in thousands of participants, so the same participant can be scored under every Allowed Information Set in a paired design. Splits follow the canonical split source and logic (\texttt{participant\_id\_split\_v1}, a participant-identifier hash allocating 0.70/0.15/0.15).

For each Task, the Scored Cohort is a seeded (20260826) uniform sample of 150 validation participants with complete demographics, CGM, and NMR evidence, drawn without reference to the target. The HbA1c Task-eligible cohort comprises 3,970 participants, of whom 607 are in the validation split and 482 have complete evidence; the corresponding counts for VAT are 8,953, 1,454, and 920. Participant identifier lists are private; SHA-256 hashes of each list and the cohort counts are published in the cohort manifests.

Each arm is an Allowed Information Set and defines one Benchmark Track for both Tasks. The demographics arm discloses age, sex, and BMI. The CGM arm adds 19 connection-level iglu summaries from the participant's primary baseline connection; the NMR arm adds 33 Nightingale analytes averaged over baseline rows; the full arm combines all three sources. A closed-book arm discloses no HPP evidence and is scored in the analysis as a constant per-model prior rather than through an Eval Run, because a Track requires at least one HPP information source. For HbA1c, a shuffled-control arm gives each participant another participant's complete packet, including demographics (no self-matches; seed 20260826), so it tests whether any of the disclosed evidence is used rather than the contribution of CGM or NMR specifically. Packet validation enforces the leakage exclusions: no participant, visit, or stage identifiers; no target field; no DXA field of any kind; a forbidden-term scan per Task; and the CGM-derived glucose management indicator and estimated A1c (\texttt{iglu\_gmi}, \texttt{iglu\_ea1c}) withheld from every LLM arm and reserved as a heuristic comparator. Units are disclosed; the Task's valid target range is not. Fasting NMR glucose is a distinct same-visit measurement and is deliberately visible in the NMR arms, to the language models and to the ridge probe alike.

The prompt contract is frozen and hashed over the system prompt, user template, closed-book prompt, and output schema (HbA1c \texttt{bf37f465...}; VAT \texttt{87f2962b...}; full hashes in the figure package). The model must return strict JSON with a numeric prediction and a free-text reasoning field. Responses that do not parse are recorded as missing predictions and counted as failures; they are never imputed. Failure rates from parsing, refusal, and transport errors are reported per arm and model.

Four External Prediction Models were queried through OpenRouter {[}29{]} with pinned model identifiers and provider routing, fallbacks disabled, and unsupported parameters rejected: \texttt{anthropic/claude-sonnet-5} (provider anthropic), \texttt{openai/gpt-5.6-luna} (openai), \texttt{google/gemini-3.7-flash} (google-ai-studio), and \texttt{deepseek/deepseek-v4-pro} (streamlake). Reasoning effort was requested as low for every model; Claude reported zero reasoning tokens in every call, so its extended thinking was not engaged, whereas the other three models emitted reasoning tokens. The output budget was 1,200 tokens, raised to 4,000 for DeepSeek only after 129 of its first-pass calls exhausted the budget on reasoning and returned no content; every DeepSeek call was then re-issued and the first-pass records were retained as superseded. No other model reached the budget. Temperature was fixed at 0 for Gemini and DeepSeek; the Claude and GPT endpoints do not accept a temperature parameter and ran at provider-default decoding. The served model, provider, token counts, cost, and generation identifier are recorded per call. Each model's predictions enter PhenoBench as an Imported Prediction Artifact bound to the Scored Cohort and its Allowed Information Set.

Baselines are scored on the same 150 participants with the same visible fields. The demographic floor follows the canonical PhenoBench baseline contract. Cross-validated ridge probes (\texttt{ridge\_cv}) fitted on the canonical training split of the Task-eligible participants who have the arm's features provide the conventional comparator for the CGM, NMR, and combined arms. For HbA1c, the raw glucose management indicator {[}30{]} used directly as the prediction provides a heuristic comparator (on this cohort it scores below the demographic floor), and each model's closed-book output provides its prior. Because the benchmark scores a pinned subset only after training on the remainder, baselines were run on the full Task-eligible validation cohort and their predictions restricted to the Scored Cohort in analysis.

The primary endpoint per Task is \(R^2\) on the Scored Cohort; mean absolute error in native units is the display metric and is shown alongside \(R^2\). Prespecified paired comparisons within each model are the grounding lift from demographics to the full packet, the modality lift from demographics to CGM and to NMR, the difference between the LLM full arm and the ridge full arm, the difference between the LLM full arm and the GMI comparator (HbA1c), and the difference between the shuffled and full arms (HbA1c), which must favor the full arm for the evidence to count as used. Uncertainty for every difference is a paired bootstrap over participants with 2,000 resamples (seed 20260826) and 95\% percentile intervals. No significance claims are made; every difference is reported with its interval, and mean signed error is reported alongside mean absolute error. Metrics are not averaged across Tasks and no post hoc composite model score is computed.

Three further External Prediction Models were queried through the Amazon Bedrock Converse API in the us-east-1 region, with pinned inference-profile identifiers: Claude Opus 4.8 (\texttt{us.anthropic.claude-opus-4-8}), Amazon Nova Pro (\texttt{us.amazon.nova-pro-v1:0}), and Llama 3.3 70B (\texttt{us.meta.llama3-3-70b-instruct-v1:0}). The output budget was 1,200 tokens for all three; temperature was fixed at 0 for Nova Pro and Llama 3.3 70B, and Claude Opus 4.8 ran at provider-default decoding. The output schema is appended to the system prompt and responses are parsed under the same strict JSON rule. Because the Converse API returns token counts but no cost field, the cost recorded per call is computed from token counts using the analysis configuration's pinned list-price assumptions (USD per million input/output tokens: 5.00/25.00 for Claude Opus 4.8, 0.80/3.20 for Nova Pro, 0.72/0.72 for Llama 3.3 70B), not a billed amount. Each Bedrock call records the same raw provenance fields as the OpenRouter client. Together with the four OpenRouter models this forms a seven-model panel; the additions were frozen before any of their results were scored.

To test ordering without requiring a calibrated number, we defined two PhenoBench-LLM group-ranking Tasks, \texttt{hba1c\_group\_rank} and \texttt{vat\_group\_rank}, which delegate target construction, eligibility, and splits to the parent Task and only partition its validation participants into groups. For each parent Task, a named cohort of validation participants with complete demographics, CGM, and NMR evidence was capped at 480 by a seeded uniform sample and partitioned into 120 groups of four by a seeded random permutation (seed 20260826), so every participant sits in exactly one group. The Evaluation Unit is the within-group pair \texttt{group:a:b}, with participants ordered by identifier, and the target is 1 if \(y_a > y_b\) and 0 otherwise; tied pairs are excluded at construction, leaving 672 of 720 possible pairs for HbA1c and all 720 for VAT. Participants in a group are labelled A to D by a seeded shuffle so that label order carries no target information. One prompt per group presents the four evidence packets under the arm's Allowed Information Set (demographics or full packet) and asks for a ranking from highest to lowest; the response must be strict JSON whose rank list is a permutation of the labels (a leading ``Participant'' before a label and code fences are tolerated), and anything else is a failure that drops the group and is counted. The system prompt reused the point-estimation wording, which refers to one participant and a single estimate; the group instruction is in the user prompt. Because a failed group removes its pairs, accuracy is reported over scored pairs together with the number of failed groups and an abstain-scored sensitivity in which every pair of a failed group receives half credit. The rank list is encoded as one hard prediction per pair (1 if \(a\) is ranked above \(b\)), submitted as a \texttt{submitted\_predictions} artifact and scored through prediction passthrough, so no new Output Channel was needed. The primary metric is pairwise accuracy, with an analytic chance level of 0.5 (a constant prediction of 0.5 receives half credit on every pair) and the mean per-group Kendall \(\tau\) between true and predicted within-group orderings as an auxiliary metric. Conventional comparators are the ridge full-packet probe and the demographic floor, whose point predictions are ordered within the same groups and pair-encoded the same way. Because pairs within a group are not independent, uncertainty for each paired difference is a group-level paired bootstrap (groups resampled with their pairs kept together; 2,000 resamples, seed 20260826) with 95\% percentile intervals. Headline ordering win rates use scored pairs and can therefore reflect different retained pair subsets; abstain-scored accuracy is reported as a sensitivity analysis.

To ask where LLM grounding helps relative to conventional models, the same regression protocol was applied across 11 further existing PhenoBench Tasks. From 78 probing continuous-scalar Tasks we excluded any target that is a packet field or a transform of one (CGM summaries, NMR analytes and their laboratory twins, eGFR, TyG, FLI, body weight, chronological age) and chose, for domain spread, Tasks with at least 337 complete-evidence validation participants: resting heart rate from the 12-lead ECG (beats per minute), thigh-to-ankle pulse wave velocity (m/s), sitting systolic blood pressure (mmHg), liver ultrasound attenuation coefficient (dB/cm/MHz), serum uric acid (mg/dL), serum TSH (mIU/L), white blood cell count (\(10^3\) cells/\(\mu\)L), apnea-hypopnea index from the overnight sleep study (events per hour), DXA femoral neck bone mineral density (g/cm\(^2\)), gut microbiome Shannon diversity (nats), and retinal fundus artery average width (AutoMorph pixels). FIB-4 was also run but is excluded from the map: age is an ingredient of the FIB-4 formula, so PhenoBench refuses its demographic floor and the conventional comparison is undefined; its calls are retained in the raw store and not analyzed. Each Task uses a seeded (20260826) sample of 150 complete-evidence validation participants with its own cohort manifest, three arms (demographics, full packet, and full packet with the Task Card) plus the closed-book prior, and four models: Gemini 3.7 Flash and GPT-5.6 via OpenRouter, Claude Opus 4.8 and Amazon Nova Pro via Bedrock. The card arm prepends to the full-packet prompt the PhenoBench Task Card rendered without its implementation facts, HPP data-source section, status lines, or any valid-range statement. The remaining card text is not purely clinical: cards quote HPP reference values (for example the cohort median and clinical thresholds for liver attenuation) and the benchmark's own baseline rows with their errors, so the card conveys the measurement's scale and typical spread as well as domain context. The card-conditioned difference therefore reflects curated HPP-specific context including reference values, not clinical knowledge alone; corresponding arm changes are also reported using the scale-free Spearman correlation between predictions and targets. For each Task and model the LLM full-packet change is \(R^2\)(full) \(-\) \(R^2\)(demographics), the card-conditioned change is \(R^2\)(card) \(-\) \(R^2\)(full), and the conventional difference is the ridge \(\Delta R^2\) = \(R^2\)(ridge full packet) \(-\) \(R^2\)(demographic floor), from validation Eval Runs using the same input fields and sampled cohort. Absolute ridge scores use all 150 participants; each LLM arm score uses the participants with a parseable estimate, at least 144 in every cell. These arm-score differences are descriptive and can involve different retained participants. The paired LLM-versus-comparator differences in the suite leaderboard instead use common evaluable participants for each comparison. Agreement between the LLM full-packet change and ridge \(\Delta R^2\) is summarized per model as Spearman \(\rho\) across Tasks with a Task-level bootstrap interval (2,000 resamples, seed 20260826). Because \(R^2\) is unbounded below, an off-scale output for a few participants produces a large negative arm \(R^2\); a large positive difference can reflect reduced scale or calibration error rather than genuine information gain. The differences are therefore interpreted alongside absolute \(R^2\), MAE in native units, and Spearman correlation. The tables report unclipped values, and the figure clips at \(\pm 0.5\) and marks clipped cells.

This evaluation used a single run, 150 participants per regression Task and 480 per ranking Task, a seven-model panel, and the validation split. It does not establish held-out performance, and if the shuffled control is not worse than the full arm for a model, evidence use is not demonstrated for that model. The aggregate leaderboard rows and Task Cards used as evidence are public; participant identifiers, evidence packets, raw requests and responses, and the Imported Prediction Artifacts with their provenance are held privately.

\subsection{PhenoBench-LLM as a benchmark suite}\label{phenobench-llm-as-a-benchmark-suite}

The evaluations above were then organised into a benchmark suite, PhenoBench-LLM, in the manner of MedHELM {[}31{]}: a taxonomy of evaluation categories laid over existing PhenoBench Tasks, with one fixed evidence packet, one scoring path, and one pair of conventional comparators per category. The suite adds no new target constructions. Every entry is either an existing participant-level PhenoBench Task at the baseline visit, a PhenoBench-LLM group-ranking Task that delegates to such a parent, or an existing longitudinal Task configured with its follow-up visit as the target. Five categories are defined: phenotype recovery (estimate a hidden continuous measurement), classification (estimate the probability of a curated binary status), longitudinal forecasting (estimate a measurement at the follow-up visit from baseline evidence), ordering (rank participants within small groups), and card-conditioned context (any of the above with the PhenoBench Task Card prepended). Task-level Eval Units, cohorts, and metrics are unchanged; only the evidence packet and the response contract are supplied by the suite.

Membership was screened over the registered participant-level Tasks against the fixed demographics, CGM, and NMR packet. A Task was excluded if its target is a packet field or a transform of one (CGM summaries, NMR analytes and their laboratory twins), if a packet field is an ingredient of the target's formula (eGFR, TyG, FLI, FIB-4, body weight, chronological age), if its label is a BMI threshold (BMI is disclosed, so the demographic floor is definitional), if its label is a curated NMR-lipid phenotype (the packet contains the lipid analytes the label is derived from), or if its prevalence is too low for positive cases to be counted reliably in a Scored Cohort of 150. Meal-level, night-level, causal-effect, and retrieval Tasks are out of scope because the packet is participant-level. Within the eligible set, Tasks were chosen for domain spread. Each Task uses a seeded (20260826) uniform sample of complete-evidence validation participants as its Scored Cohort, with its own cohort manifest carrying counts and identifier-list hashes, and the same leakage validator, forbidden-term scan, and strict-JSON parsing rule as the point-estimation protocol.

Phenotype recovery comprises the two anchor Tasks and the eleven cross-Task map Tasks described above together with twelve further Tasks: sitting diastolic blood pressure (mmHg), QT interval from the 12-lead ECG (ms), mean carotid intima-media thickness (mm), liver stiffness by ultrasound elastography (kPa), DXA lumbar spine L1--L4 bone mineral density (g/cm\(^2\)), maximum hand grip strength of the better hand (kg), blood hemoglobin (g/dL), serum ferritin (ng/mL), serum alanine aminotransferase (U/L), minimum nocturnal oxygen saturation from the overnight sleep study (\%), sleep efficiency (\%), and the minimum ankle-brachial index over both sides (ratio). The primary metric is \(R^2\) with mean absolute error in native units alongside, as before. Classification comprises five curated HPP phenotypes, each a participant-level binary label built from measurements, diagnoses, questionnaires, or medication rather than from a threshold on any packet field: hypertension, metabolic-associated fatty liver disease, prediabetes or diabetes, hypothyroidism, and migraine. Irritable bowel syndrome was also screened and evaluated but was excluded from the reported suite; its records are retained in the raw store and excluded from tables and figures. The prediabetes Task is flagged as a glycaemic-proxy Task: NMR fasting glucose, a laboratory twin of one of the label's criteria, is visible in the packet, so it tests a proxy rather than a clean curated phenotype. Because the prevalence of hypothyroidism and of migraine in the complete-evidence validation pool is low, their Scored Cohorts are seeded uniform samples of 300 rather than 150 participants, so that the expected number of positive cases supports a rank-based metric. Longitudinal forecasting comprises four Tasks whose parents are already in the suite: sitting systolic and diastolic blood pressure (mmHg), liver ultrasound attenuation coefficient (dB/cm/MHz), and apnea-hypopnea index (events per hour), each with the target measured at the follow-up visit about two years after baseline. Ordering comprises six PhenoBench-LLM group-ranking Tasks with the protocol described above: the two described earlier (HbA1c and VAT) and four whose parents were admitted by the same rule (a demographic floor \(R^2\) between 0.05 and 0.7 and at least 480 complete-evidence validation participants): sitting systolic blood pressure, apnea-hypopnea index, mean carotid intima-media thickness, and femoral neck bone mineral density. Further parent Tasks that satisfy the rule are listed in the taxonomy as candidates. The card-conditioned category re-uses the Tasks of the other categories and adds no cohort of its own.

For classification Tasks the model is asked for the probability, between 0 and 1, that the participant carries the curated status as determined at the same study visit; the output schema bounds the number to \([0, 1]\), and a response outside the bound or not parseable as strict JSON is a failure. Probabilities enter PhenoBench through the same participant-prediction passthrough as the continuous Tasks and are scored by the benchmark's binary-classification path. The primary metric is the area under the receiver operating characteristic curve (AUROC), which does not depend on the calibration of the returned probabilities; the Brier score is the secondary metric and does. The floor is the canonical PhenoBench demographic classifier floor, a logistic regression on age, sex, and BMI fitted on the Task's training split. The conventional comparator is a logistic regression fitted on the training split over exactly the fields visible in the arm (demographics with CGM, with NMR, or with both), returning probabilities, so that the comparator for a classification Task plays the role the ridge probe plays for a continuous Task. As for the continuous Tasks, baselines are run on the full Task-eligible validation cohort and restricted to the Scored Cohort in analysis.

Longitudinal Tasks are existing PhenoBench scalar Tasks configured with the follow-up visit as the target stage and the baseline visit as the covariate stage; eligibility requires a finite baseline value of the target, and the Scored Cohort is drawn from validation participants with complete baseline evidence and a follow-up measurement. The evidence packet is the baseline packet of the parent Task with one additional block: the baseline value of the target itself, in the same unit. This block is disclosed in every arm, to the language models and to the ridge comparator alike, and the leakage validator admits it for follow-up Tasks only; the prompt states that all evidence, including the baseline value of the same measurement, was measured at the baseline visit. The question is therefore whether the model can use the baseline packet to anticipate change over roughly two years, not whether it can reproduce a value it has already been given. Scoring is \(R^2\) on the follow-up value, with the demographic floor, the ridge probe over the same fields (baseline value included), and a baseline-carry-forward comparator that predicts the baseline value itself as comparators. Because the follow-up value is dominated by its autocorrelation with the baseline, \(R^2\) on the follow-up value is high for any estimate close to the disclosed baseline, and the paired comparison with the carry-forward comparator is the informative one; \(R^2\) on the change score was not prespecified and is not reported.

The External Prediction Model panel of the suite comprises fourteen models: the seven described above, two further Bedrock models, and five further OpenRouter models. Pixtral Large 2502 (\texttt{us.mistral.pixtral-large-2502-v1:0}, temperature 0) and Claude Haiku 4.5 (\texttt{us.anthropic.claude-haiku-4-5-20251001-v1:0}, provider-default decoding) were added through the Amazon Bedrock Converse API in us-east-1 with a 1,200-token output budget and cost computed from token counts and pinned list prices (USD per million input/output tokens: 2.00/6.00 and 1.00/5.00 respectively). Amazon Nova Premier and the Bedrock Llama 4 profile were also requested but the account was denied access to both inference profiles, so neither was called. The OpenRouter additions were chosen for spread across developers, model sizes, and release dates among models that support a JSON output schema, and each is pinned to a single provider with fallbacks disabled, as before: Gemini 3.1 Pro (\texttt{google/gemini-3.1-pro-preview}, google-ai-studio), GPT-5.4 nano (\texttt{openai/gpt-5.4-nano}, openai), Gemini 3.1 Flash-Lite (\texttt{google/gemini-3.1-flash-lite}, google-ai-studio), Llama 4 Maverick (\texttt{meta-llama/llama-4-maverick}, deepinfra), and Grok 4.3 (\texttt{x-ai/grok-4.3}, xai). Temperature was fixed at 0 wherever the endpoint accepts it; the GPT-5.4 nano endpoint does not and ran at provider-default decoding. Reasoning effort was requested as low, except minimal for GPT-5.4 nano and none for Llama 4 Maverick, which does not emit reasoning tokens. The output budget was 1,200 tokens, set to 4,000 from the outset for Grok 4.3 and the reasoning-emitting open-weight models, which emit reasoning tokens that count against the budget, so that the truncation seen for DeepSeek under the smaller budget could not recur. Because per-call cost scales with the number of Tasks, the two most expensive additions, Gemini 3.1 Pro and Grok 4.3, were run on a prespecified subset only: the thirteen point-estimation Tasks of the evaluations described above, three classification Tasks, and the six ranking Tasks. Gemini 3.1 Pro and Grok 4.3 each covered 22 of the 40 task formulations; in the final retained records, Claude Sonnet 5 and DeepSeek V4 Pro each covered 29, and the other ten models covered all 40. The 22-Task subset was fixed before either model's results were scored, and the number of Tasks each model was run on is reported alongside every summary of it. Each addition was frozen before any of its results were scored. Three further open-weight models, Qwen 3.8-27B, Mistral Large 2512, and GLM-5.2, were started and withdrawn before completing the suite because their per-call latency made them the limiting step; their partial records are retained but excluded from every table and figure.

Two prompt components were revised for the suite, and both revisions are versioned so that their raw records never collide with the earlier ones. The Task Card renderer (v2) keeps the card's clinical sections and, in addition to dropping the HPP data-source section, implementation facts, status lines, and any valid-range statement as before, drops every table, every section whose heading refers to the benchmark, its baselines, variants, final-test or leaderboard results, and every line that quotes a number with a unit or a summary statistic (medians, means, standard deviations, error metrics, thresholds, percentiles, or sample sizes). The v2 card therefore conveys the measurement's clinical meaning and the reasons it matters without the reference values that let the v1 card fix the target's scale, and the card lift under v2 is interpretable as the value of curated domain context rather than of disclosed reference values. The grouped-ranking system prompt (v2) instructs the model that it is comparing several participants and must return their ordering, replacing the re-used single-participant point-estimation system prompt of the first ranking run; the user prompt and output schema are unchanged. The card and ranking arms in the suite use the v2 components. The v1 card-arm results reported above are retained as sensitivity rows in the tables and are not re-scored; where the same Task appears under both versions, the two are shown side by side.

The suite is summarised in a leaderboard with two levels. The per-Task level has one row per Task and model: the model's full-packet score on the category's primary metric (or, for ordering, its pairwise accuracy), the demographic floor and the conventional probe (for ordering, the floor-ordering and ridge-ordering comparators), the paired differences of the model from each, and the model's demographics-only and card-conditioned scores where those arms were run. Absolute arm scores use each arm's valid outputs. Differences against the demographic floor and fitted probe use common evaluable participants or ranking groups for each paired comparison; the fitted probe receives the same input fields as the LLM. The follow-up fraction beating baseline carry-forward likewise uses paired estimates on common evaluable participants. These differences condition on valid outputs and do not count missing predictions as incorrect predictions. The category level summarises, for each category and model, the number of Tasks, the share of Tasks on which the model's full arm exceeds the floor and the share on which it exceeds the conventional probe, the mean and median of the paired differences, and the model's mean rank among the language models across the category's Tasks (rank 1 is best, ties averaged). Ranks are computed within a Task, so they compare models to one another and not across Tasks. No composite score is formed across categories, because the primary metrics (\(R^2\), AUROC, pairwise accuracy) are not commensurable and the Tasks are not a probability sample of anything. The leaderboard is descriptive: it uses a single run per cell on the validation split with a fixed evidence packet, its per-Task differences carry the paired bootstrap intervals described earlier, and its category summaries do not support general model rankings. It claims where, on these Tasks and this evidence, a model's full-packet estimate does or does not exceed the demographic floor and a conventional probe fitted on the same fields; it does not claim held-out performance, a ranking of models in general, or that the categories are exhaustive of the questions PhenoBench can ask. Three further leaderboard statistics compare the models with one another rather than with the comparators. The win rate, adapted from MedHELM {[}31{]}, is computed over all pairs of models: for every Task on which both models of a pair have a full-packet score, the model with the higher primary metric wins the head-to-head and a tie counts half for each; a model's win rate is the share of its head-to-heads won. Its 95\% interval is a percentile interval from a Task-level bootstrap (Tasks resampled with replacement, 2,000 resamples, seed 20260826), so it reflects sensitivity to the Task set rather than to participants. Because a head-to-head requires both models to have been run on the Task, the win rate is reported at two levels: the headline win rate for cross-model comparison is computed on the common Task set, the Tasks that every retained model ran, and a secondary win rate is computed over all available head-to-heads, with each model's Task count alongside; on the secondary rate a model run on a subset of Tasks is compared only on that subset and is comparable with the full-suite models descriptively, not inferentially. Each leaderboard cell (one Task, one model, full-packet arm) carries a participant-level bootstrap interval of its primary metric (\(R^2\) or AUROC; 1,000 resamples, seed 20260826, 95\% percentile), computed over the participants for whom the model returned a parseable estimate. At Scored Cohorts of 150 or 300 participants the AUROC intervals are wide (0.11--0.35), so differences between models, and between a model and the logistic comparator, on the classification Tasks are descriptive only; where a fitted comparator scores below the demographic floor on the Scored Cohort, as the logistic comparator does for migraine, the floor is the operative comparator for that Task. One Task, retinal fundus artery average width, exposes a unit-scale failure shared by every model: the target is in AutoMorph pixels and every model answers on a micrometre-like scale, so every estimate is off-scale; the Task is retained in the tables as a contract-following failure. Cost per call is the mean over the model's non-errored calls of the recorded cost, which is the OpenRouter-reported amount for OpenRouter models and the token counts priced at the pinned list prices for Bedrock models, together with mean prompt and completion token counts. Two failure rates are reported per model across its Tasks: the transport-error rate, the share of calls that returned no response, and the parse-failure rate, the share of returned responses that were not strict JSON. A prespecified failure-mode table, computed from the full-packet arm of each continuous Task, reports per Task and model the abstention rate (the share of requested participants without a parseable estimate), the off-scale rate (the share of estimates further than three cohort standard deviations from the cohort mean of the target), the bias (mean signed error in units of the cohort standard deviation), and the spread ratio (the standard deviation of the estimates divided by that of the target), so that a poor \(R^2\) can be attributed to a few off-scale outputs, a systematic offset, or compressed predictions.

The main figure separates capability profile from efficiency under the fixed cohort-data contract. Panel A reports each model's within-category pairwise win rate using only same-Task head-to-heads, with unequal category coverage retained explicitly. Panel B reports common-Task pairwise win rate against mean cost per call and identifies the Pareto frontier. The evaluation contract and representative question, conventional-probe references, absolute per-Task estimates, paired evidence analyses, Task Card effects, failure modes, and reliability details are reported in the text and Supplementary Figures. The taxonomy, cohort-level manifests, evaluation settings, leaderboard tables, and metric code are public.

\section{Extended limitations}\label{extended-limitations}

The question set is broad but uneven and reflects the questions formulated and implemented so far. Some empirical comparisons jointly change the measurement and its representation, so they estimate the value of the complete route. The screening analyses span many questions without multiplicity adjustment, and their predictive associations do not establish causality or clinical utility. HPP-specific recruitment, measurement protocols and missingness may also limit transfer to other cohorts.

Subgroup robustness and fairness remain open questions. We have not assessed how aggregate model comparisons change with clinical-domain composition.

The model comparisons cover only tasks compatible with each method's input and output requirements. The 160 complete-case cells are nested within 52 tasks and are not independent clinical questions; this post hoc cross-task synthesis was not preregistered. Whole-task resampling and equal-task aggregation address repeated tracks within a task, but tasks still share participants and related outcomes. For non-ridge methods, \(\Delta R^2\) relative to the common ridge-fitted demographic baseline combines differences in fitting demographics with differences in using the added predictors, and uncertainty does not resample model fitting or establish independent population replication. In the same-signal analysis, the exact-intersection WatchPAT rerun was post hoc, ECG lacked an engineered comparator, and only a ridge head was tested on embeddings with 770--3,076 dimensions; no tabular foundation model was evaluated on them. Cohort matching makes the practical ridge-head comparison valid, but the dimensional imbalance prevents interpreting it as an intrinsic ranking of representation quality. The machine-querying experiment is a proof of concept; repeated epochs reused the same questions, and results depend on the tested prompts, tools, harness, and provider implementations. Neither analysis supports a general model ranking or clinical utility.

The cohort-grounded language-model evaluation used one run per cell on validation cohorts of 150--480 participants, a fixed packet of summary features and one prompt family. Category estimates retain unequal Task coverage, common-Task win rates summarize sensitivity to the tested Task set, and cost estimates combine provider-reported amounts with token counts priced from pinned list prices. The results do not establish held-out performance, general model superiority or clinical utility. Training-data exposure to published HPP findings or reference distributions cannot be ruled out for the evaluated LLMs. Withholding participant-level evaluation data does not establish absence of such exposure; published population summaries and individual-level target leakage are distinct risks. The shuffled-packet control tests participant-specific information use in its evaluated setting, not absence of pretraining contamination. The comparison with cohort-fitted probes also combines differences in training access, calibration and model behavior, rather than isolating biomedical reasoning.

A participant-level holdout prevents direct overlap between training and test participants within an analysis, but it does not preserve independence under repeated use of the same cohort. Later choices of outcomes, features and models can be informed by earlier results even when every analysis uses a clean split {[}8,10{]}. This concern is familiar in large biobanks, where discovery and replication may share data-collection and processing dependencies {[}9{]}, and reporting may not establish whether evidence was independently replicated {[}11{]}. Expansion therefore increases the need for prospective task definitions, versioned results, protected test tasks and independent-cohort validation. PhenoBench provides standardized internal evidence; external validation remains necessary for claims intended to generalize beyond HPP.

\section{LLM usage disclosure}\label{llm-usage-disclosure}

Large language models were evaluated as part of this study. Generative AI tools were also used for agent-assisted task curation, creating and editing software and analysis code, literature and citation support, drafting and editing manuscript text including Methods and appendices, and verification. Figure 1 includes AI-assisted conceptual artwork. Its labels and schematic layout were checked against the benchmark specification. The authors reviewed all AI-assisted outputs, checked analyses and citations against their source materials, and read, tested where applicable, and approved the resulting text and artifacts. The authors take responsibility for the final content.

\section{Registered task inventory}\label{registered-task-inventory}

\label{sec:task-inventory}

The following index lists all 90 registered clinical targets. It excludes three framework-only diagnostics. The corresponding Task Cards describe target construction, clinical context, timing and documented eligibility caveats. Each catalog Track additionally records the target/unit, allowed inputs and scored split. A task can have multiple Tracks with different cohorts and inputs, so it has no single sample count or baseline performance.

The ten-method regression grid includes 52 registered targets; 30 registered targets underlie the 40 LLM formulations: 25 phenotype-recovery (P), 5 classification (C), 4 follow-up (F) and 6 ordering (O) formulations. Sixteen targets occur in both headline analyses, and 24 occur in neither. The counts describe analysis coverage, not a partition of the registry. Tab denotes inclusion in the 160-cell grid; a dash means absence from that analysis.

Fifteen targets have no result row in the released catalog, including 14 with no released Track. NR denotes that absence, not zero performance or evidence that a target has never been evaluated. Three of these targets occur in the separate LLM suite: VAT and femoral-neck BMD use 150 participants for phenotype recovery and 480 for ordering; lumbar-spine BMD uses 150 for phenotype recovery.

Catalog \(N\) and \(n\) give the total and scored counts across the result rows in the versioned public catalog. These ranges only describe those rows. For example, the earlier meal-trajectory evaluation in Figure 2A and the fasting-glucose validation runs in Figure 2 use the separate cohorts described in Appendix A.2. Counts for meal and participant-night tasks are in meals and participant-nights. Class prevalences and formal per-card adjudication records are not tabulated in the released inventory. Matched regression rows separately retain their demographic-ridge baseline.

The versioned benchmark catalog provides the task IDs and Track-level records underlying this index. Complete eligibility expressions are not tabulated for every target. Shared Task Cards distinguish the clinical and NMR lipid targets and the two FIB scores.

\noindent\textbf{Registered clinical targets and coverage.} Catalog counts are not matched across rows. ATE: average treatment effect; R@1: Recall@1; BA: balanced accuracy.

\begingroup
\ifdefined\nolinenumbers\nolinenumbers\fi
\footnotesize
\setlength{\tabcolsep}{2pt}
\renewcommand{\arraystretch}{1.10}
\begin{longtable}{@{}>{\raggedright\arraybackslash}p{0.31\textwidth}>{\raggedright\arraybackslash}p{0.08\textwidth}>{\raggedright\arraybackslash}p{0.17\textwidth}>{\raggedright\arraybackslash}p{0.16\textwidth}>{\raggedright\arraybackslash}p{0.07\textwidth}>{\raggedright\arraybackslash}p{0.09\textwidth}@{}}
\toprule
Registered target & Metric & Catalog $N$ & Catalog $n$ & Tab & LLM \\
\midrule
\endfirsthead
\toprule
Registered target & Metric & Catalog $N$ & Catalog $n$ & Tab & LLM \\
\midrule
\endhead
\midrule
\multicolumn{6}{r}{Continued on next page}\\
\endfoot
\bottomrule
\endlastfoot
\multicolumn{6}{@{}l}{\textbf{Hematologic}}\\
Ferritin & $R^2$ & 1,778--3,671 & 257--559 & Yes & P \\
Hemoglobin & $R^2$ & 1,448--4,502 & 228--687 & Yes & P \\
\multicolumn{6}{@{}l}{\textbf{Immune}}\\
Absolute Basophil Count & $R^2$ & 1,448--4,857 & 228--749 & Yes & --- \\
Absolute Eosinophil Count & $R^2$ & 1,448--4,857 & 228--749 & Yes & --- \\
Absolute Lymphocyte Count & $R^2$ & 1,448--4,857 & 228--749 & Yes & --- \\
Absolute Monocyte Count & $R^2$ & 1,448--4,857 & 228--749 & Yes & --- \\
Absolute Neutrophil Count & $R^2$ & 1,448--4,857 & 228--749 & Yes & --- \\
Basophil Percentage & $R^2$ & NR & NR & --- & --- \\
Eosinophil Percentage & $R^2$ & NR & NR & --- & --- \\
Lymphocyte Percentage & $R^2$ & NR & NR & --- & --- \\
Monocyte Percentage & $R^2$ & NR & NR & --- & --- \\
Neutrophil Percentage & $R^2$ & NR & NR & --- & --- \\
White blood cell count & $R^2$ & 1,448--4,857 & 228--749 & Yes & P \\
\multicolumn{6}{@{}l}{\textbf{Cardiovascular and lipid}}\\
ApoB & $R^2$ & 6,574--9,134 & 1,006--1,400 & Yes & --- \\
Carotid intima-media thickness & $R^2$ & 6,983--7,403 & 1,057--1,123 & Yes & P/O \\
Diastolic blood pressure & $R^2$ & 2,929--9,426 & 460--1,425 & Yes & P/F \\
HDL cholesterol (NMR) & $R^2$ & 1,900--9,134 & 289--1,400 & Yes & --- \\
Hyperlipidemia status & AUROC & 2,358--3,495 & 334--504 & --- & --- \\
Hypertension status & AUROC & 4,025--5,125 & 633--785 & --- & C \\
LDL cholesterol & $R^2$ & 1,892--3,152 & 288--483 & Yes & --- \\
LDL cholesterol (NMR) & $R^2$ & 6,573--9,132 & 1,006--1,400 & Yes & --- \\
Mean ankle pressure & $R^2$ & 3,328--8,669 & 516--1,308 & Yes & --- \\
Minimum bilateral ABI & $R^2$ & 3,328--8,669 & 516--1,308 & Yes & P \\
Non-HDL cholesterol & $R^2$ & 740--5,430 & 100--819 & Yes & --- \\
1-minute orthostatic SBP drop & $R^2$ & 1,575--8,466 & 239--1,297 & Yes & --- \\
PR interval & $R^2$ & 5,495--9,330 & 853--1,421 & Yes & --- \\
QRS duration & $R^2$ & 5,578--9,414 & 860--1,432 & Yes & --- \\
QT interval & $R^2$ & 5,577--9,410 & 860--1,432 & Yes & P \\
QTc & $R^2$ & 5,576--9,409 & 860--1,432 & Yes & --- \\
Resting heart rate & $R^2$ & 9,382--9,420 & 1,419--1,434 & Yes & P \\
Systolic blood pressure & $R^2$ & 7,803--9,425 & 1,192--1,425 & Yes & P/F/O \\
Triglycerides & $R^2$ & 3,573--5,751 & 535--865 & Yes & --- \\
Triglycerides (NMR) & $R^2$ & 6,574--9,134 & 1,006--1,400 & Yes & --- \\
Thigh-ankle pulse-wave velocity & $R^2$ & 5,911--7,592 & 918--1,136 & Yes & P \\
\multicolumn{6}{@{}l}{\textbf{Endocrine}}\\
Hypothyroidism Status & AUROC & 9,001--10,128 & 1,351--1,542 & --- & C \\
TSH & $R^2$ & 2,390--4,707 & 360--711 & Yes & P \\
Vitamin D & $R^2$ & 1,268--2,497 & 189--374 & Yes & --- \\
\multicolumn{6}{@{}l}{\textbf{Metabolic and metabolomics}}\\
CGM coefficient of variation & $R^2$ & 7,722--9,064 & 1,171--1,385 & Yes & --- \\
CGM mean glucose & $R^2$ & 7,722--9,064 & 1,171--1,385 & Yes & --- \\
CGM time in range 70--180 mg/dL & $R^2$ & 7,722--9,064 & 1,171--1,385 & Yes & --- \\
Fasting plasma glucose & $R^2$ & 3,423--6,843 & 495--1,029 & Yes & --- \\
GlycA & $R^2$ & 2,760--9,147 & 439--1,403 & Yes & --- \\
Glycine (NMR) & $R^2$ & 4,613--9,149 & 697--1,403 & Yes & --- \\
HbA1c & $R^2$ & 304--3,669 & 45--549 & --- & P/O \\
Meal-conditioned CGM trajectory & RMSE & 231,207--258,421 & 35,175--39,394 & --- & --- \\
Meal postprandial glucose response & $R^2$ & 258,421 & 39,394 & --- & --- \\
Prediabetes Status & AUROC & 8,868--11,317 & 1,329--1,720 & --- & C \\
T2D Status & AUROC & 4,243 & 619 & --- & --- \\
Total BCAA (NMR) & $R^2$ & 4,613--9,149 & 697--1,403 & Yes & --- \\
Triglyceride--glucose index & $R^2$ & 2,831--5,651 & 403--848 & Yes & --- \\
\multicolumn{6}{@{}l}{\textbf{Renal and metabolic}}\\
eGFR & $R^2$ & NR & NR & --- & --- \\
Urate & $R^2$ & 537--2,250 & 81--332 & --- & P \\
\multicolumn{6}{@{}l}{\textbf{Hepatobiliary}}\\
Albumin & $R^2$ & 2,154--4,499 & 305--672 & Yes & --- \\
ALT & $R^2$ & 2,961--6,051 & 435--917 & Yes & P \\
Fatty Liver Disease Status & AUROC & 6,025--7,207 & 920--1,083 & --- & C \\
FIB-3 & $R^2$ & 2,307--4,786 & 338--730 & Yes & --- \\
FIB-4 & $R^2$ & 2,907--4,371 & 443--656 & Yes & --- \\
Fatty liver index & $R^2$ & 1,270--1,902 & 179--269 & Yes & --- \\
Liver Attenuation & $R^2$ & 5,024--8,594 & 748--1,284 & Yes & P/F \\
Liver Stiffness & $R^2$ & 5,025--8,600 & 747--1,286 & Yes & P \\
\multicolumn{6}{@{}l}{\textbf{Anthropometry and body composition}}\\
Body Weight & $R^2$ & NR & NR & --- & --- \\
Lean Body Mass & $R^2$ & 4,695--7,817 & 708--1,186 & Yes & --- \\
Obesity Status & AUROC & NR & NR & --- & --- \\
Visceral adipose tissue mass & $R^2$ & NR & NR & --- & P/O \\
Waist & $R^2$ & NR & NR & --- & --- \\
Waist Hip Ratio & $R^2$ & NR & NR & --- & --- \\
\multicolumn{6}{@{}l}{\textbf{Sleep physiology}}\\
Diet Sleep Causal Effect & ATE & NR & NR & --- & --- \\
Next Night Sleep & $R^2$ & 18,242--29,642 & 2,779--4,578 & --- & --- \\
Sleep Efficiency & $R^2$ & 2,155--4,020 & 335--625 & --- & P \\
Total Sleep Time & $R^2$ & 2,147--4,015 & 332--625 & --- & --- \\
\multicolumn{6}{@{}l}{\textbf{Respiratory and hypoxaemia}}\\
Apnoea--hypopnoea index & $R^2$ & 7,335--8,180 & 1,121--1,231 & --- & P/F/O \\
Asthma Status & AUROC & 7,130--8,418 & 1,080--1,292 & --- & --- \\
Minimum nocturnal SpO2 & $R^2$ & 3,625--8,180 & 565--1,231 & --- & P \\
Oxygen desaturation index & $R^2$ & 3,625--8,180 & 565--1,231 & --- & --- \\
\multicolumn{6}{@{}l}{\textbf{Gut and microbiome}}\\
Gut Microbiome Identity Retrieval & R@1 & NR & NR & --- & --- \\
Gut Shannon Diversity & $R^2$ & 2,475--4,706 & 378--727 & --- & P \\
Gut Species Richness & $R^2$ & 2,475--4,706 & 378--727 & --- & --- \\
Irritable Bowel Syndrome Status & AUROC & 5,046--11,726 & 778--1,783 & --- & --- \\
\multicolumn{6}{@{}l}{\textbf{Ophthalmic and retinal}}\\
Fundus Artery Average Width & $R^2$ & 3,016--6,775 & 463--1,038 & Yes & P \\
Fundus Artery Fractal Dimension & $R^2$ & 3,016--6,775 & 463--1,038 & Yes & --- \\
Fundus Artery Tortuosity Density & $R^2$ & 3,016--6,775 & 463--1,038 & Yes & --- \\
Fundus Artery Vessel Density & $R^2$ & 3,016--6,775 & 463--1,038 & Yes & --- \\
Fundus Vein Average Width & $R^2$ & 3,016--6,775 & 463--1,038 & Yes & --- \\
Fundus Vein Tortuosity Density & $R^2$ & 3,016--6,775 & 463--1,038 & Yes & --- \\
\multicolumn{6}{@{}l}{\textbf{Musculoskeletal}}\\
Femoral neck BMD & $R^2$ & NR & NR & --- & P/O \\
Lumbar spine BMD & $R^2$ & NR & NR & --- & P \\
Maximum Bilateral Grip Strength & $R^2$ & 5,111--11,404 & 779--1,731 & Yes & P \\
\multicolumn{6}{@{}l}{\textbf{Neurologic and psychosocial}}\\
Loneliness Status & BA & 6,493 & 966 & --- & --- \\
Migraine Status & AUROC & 8,167--9,057 & 1,243--1,378 & --- & C \\
\multicolumn{6}{@{}l}{\textbf{Aging}}\\
Chronological Age & $R^2$ & 5,727--11,483 & 862--1,745 & --- & --- \\
\end{longtable}
\endgroup
\end{document}